\documentclass[10pt,twocolumn,letterpaper]{article}

\usepackage[pagenumbers]{cvpr} % To force page numbers, e.g. for an arXiv version
\usepackage[utf8]{inputenc}
\usepackage[T1]{fontenc}
\usepackage{url}
\usepackage{booktabs}
\usepackage{amsfonts}
\usepackage{nicefrac}
\usepackage{microtype}
\usepackage{graphicx}
\usepackage{subcaption}
\graphicspath{{media/}}
\usepackage{tabularx}
\usepackage{adjustbox}
\usepackage{pgfplots}
\usepackage[ruled,vlined]{algorithm2e}
\usepackage{amsmath}
\usepackage{amssymb}
\usepackage{amsthm}
\usepackage{float}
\usepackage{setspace}
\usepackage{xcolor}
\pgfplotsset{compat=1.18}
\usepackage{enumitem}
\usepackage{multirow}
\usepackage{siunitx}
\usepackage{xfrac}
\usepackage{algorithmic}

\renewcommand{\paragraph}[1]{\vspace{1.25mm}\noindent\textbf{#1}}

\DeclareMathOperator*{\argmax}{arg\,max}

\newlength\mylen
\newcommand\myinput[1]{%
  \settowidth\mylen{\KwIn{}}%
  \setlength\hangindent{\mylen}%
  \hspace*{\mylen}#1\\}

\newcommand{\red}[1]{{\color{red}#1}}
\newcommand{\blue}[1]{\textcolor{blue}{#1}}

\definecolor{cvprblue}{rgb}{0.21,0.49,0.74}
\usepackage[pagebackref,breaklinks,colorlinks,allcolors=cvprblue]{hyperref}

\def\paperID{} % *** Enter the Paper ID here
\def\confName{CVPR}
\def\confYear{2026}

\title{ATAC: Augmentation-Based Test-Time Adversarial Correction for CLIP}

\author{
Linxiang Su \qquad András Balogh\\
University of Szeged\\
{\tt\small su\_linxiang@126.com} \qquad {\tt\small abalogh@inf.u-szeged.hu}
}

\begin{document}
\maketitle

\begin{abstract}
Despite its remarkable success in zero-shot image–text matching, CLIP remains highly vulnerable to adversarial perturbations on images. As adversarial fine-tuning is prohibitively costly, recent works explore various test-time defense strategies; however, these approaches still exhibit limited robustness. In this work, we revisit this problem and propose a simple yet effective strategy: Augmentation-based Test-time Adversarial Correction (ATAC). Our method operates directly in the embedding space of CLIP, calculating augmentation-induced drift vectors to infer a semantic recovery direction and correcting the embedding based on the angular consistency of these latent drifts. Across a wide range of benchmarks, ATAC consistently achieves remarkably high robustness, surpassing that of previous state-of-the-art methods by nearly 50\% on average, all while requiring minimal computational overhead. Furthermore, ATAC retains state-of-the-art robustness in unconventional and extreme settings and even achieves nontrivial robustness against adaptive attacks. Our results demonstrate that ATAC is an efficient method in a novel paradigm for test-time adversarial defenses in the embedding space of CLIP.
\end{abstract}

\section{Introduction}

Vision–language models (VLMs) trained on web-scale image–text corpora have transformed zero-shot recognition and open-world retrieval, with CLIP emerging as a widely adopted foundation model for image–text alignment~\citep{Radford2021LearningTV}. As such models migrate to safety- and security-critical applications, robustness becomes indispensable: small, human-imperceptible perturbations can reliably induce arbitrary errors in neural networks~\citep{ilyas2019adversarialexamplesbugsfeatures,goodfellow2015explainingharnessingadversarialexamples,carlini2017evaluatingrobustnessneuralnetworks,andriushchenko2020squareattackqueryefficientblackbox}, and CLIP is no exception. Its vulnerability raises concerns about trustworthiness and deployment risks, given its growing influence on machine perception and visual reasoning pipelines~\citep{ma2025safetyscalecomprehensivesurvey,zhou2024revisiting,mao2023understanding}.

% Pipeline included here so that it shows up on the first page
\begin{figure}
    \centering
    \includegraphics[width=\linewidth]{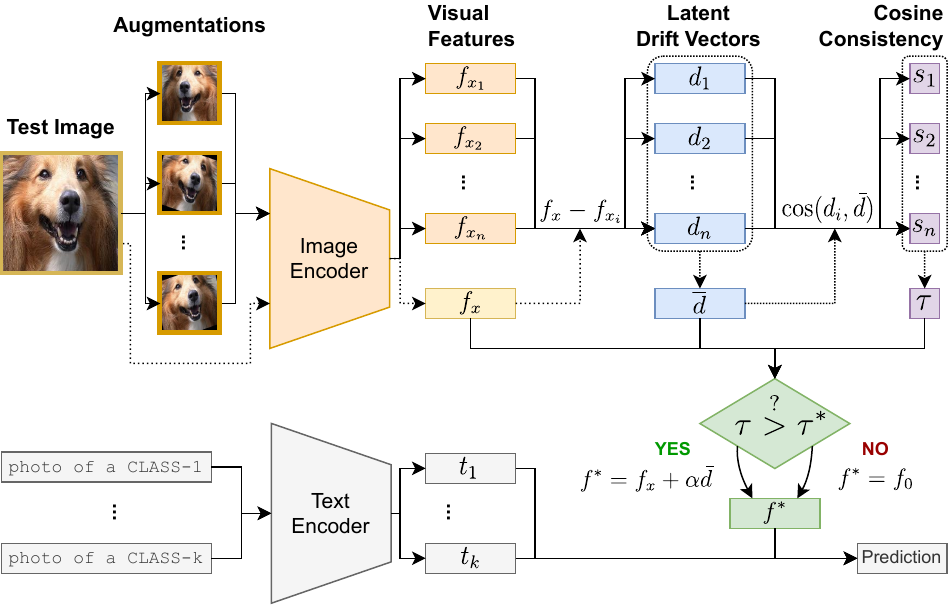}
    \caption{Overview of the ATAC framework. We use the visual features of augmented views to estimate a semantic recovery direction $\bar{d}$, and a cosine consistency gate to control the correction of the visual embedding.}
    \label{fig:atac_overview}
\end{figure}

A rich body of research has sought to improve robustness from three complementary directions. Training-time methods (e.g., adversarial training, TRADES)~\citep{Madry2017TowardsDL,schlarmann2024robustclip,mao2023understanding} offer stronger worst-case guarantees but are prohibitively costly at the scale of foundation models, and may erode zero-shot generalization~\citep{xing2025clipstrongfightback}. Input-space purification and randomized smoothing~\citep{guo2018counteringadversarialimagesusing,8578289,nie2022diffusionmodelsadversarialpurification,cohen2019certifiedadversarialrobustnessrandomized} reduce attack effectiveness but often trade off clean accuracy and remain susceptible to adaptive attacks. In contrast, \emph{test-time} strategies adapt inference-time behavior without retraining.
For CLIP, two main directions have been explored: \emph{prompt-side adaptation} (e.g., test-time prompt tuning) adjusts textual embeddings to counter undesirable and adversarial distribution shifts~\citep{shu2022testtimeprompttuningzeroshot,sheng2025rtptimprovingadversarialrobustness}, while \emph{image-side counterattacks} refine inputs at inference to push predictions toward ground-truth classes~\citep{xing2025clipstrongfightback}. Despite rapid progress, both approaches exhibit limitations: gradient-based counterattacks are computationally expensive and rely on sensitive hyperparameters, while prompt-only tuning relies on the unstable assumption that the predictions of augmented views can be aggregated to counter adversarial attacks.
%do not guarantee semantically meaningful corrections in CLIP’s feature space, while prompt-only tuning assumes that augmented logits can realign under stronger attacks--a condition that fails when residual adversarial corruption persists.

We take a different route grounded in empirical observations: although adversarial perturbations can flip CLIP’s zero-shot prediction, CLIP’s \emph{embeddings} remain comparatively stable under standard augmentations (e.g., flip, rotation, color jitter)~\cite{dahal2025embeddingshiftdissectionclip}. We demonstrate that adversarially perturbed samples produce \emph{aligned} augmentation-induced shifts in CLIP's embedding space, while clean samples exhibit \emph{scattered} shifts. Building on this, we propose \textbf{A}ugmentation-based \textbf{T}est-time \textbf{A}dversarial \textbf{C}orrection (ATAC), a simple yet powerful test-time defense that operates directly in CLIP’s embedding space.
ATAC constructs multiple augmented views, computes their embedding \emph{drift vectors} relative to the original, and averages them to estimate a semantic recovery direction. The original embedding is then corrected along this direction by a step $\alpha$, gated by a cosine-consistency threshold $\tau$ that suppresses unnecessary corrections for clean inputs.%updates for likely clean inputs.
Conceptually, ATAC bridges the gap between prompt-side tuning and input-space optimization by performing a lightweight, training-free, feature-level correction with negligible overhead.

Across 13 classification benchmarks, ATAC consistently yields remarkable improvements over both test-time defenses and CLIP-specific adversarial fine-tuning methods. Our method improves previous state-of-the-art robustness benchmarks by nearly 50\% on average with minimal computational overhead. Further analysis reveals that ATAC is able to effectively exploit a fundamental flaw in untargeted, gradient-based attacks; however, our method also achieves state-of-the-art results when evaluated against attacks that limit or eliminate this flaw. Moreover, ATAC is effective against adaptive attacks from both the perspective of robustness and computational cost. Our results highlight a new paradigm for test-time adversarial defense: \emph{direct semantic correction in CLIP’s embedding space}.
\\
\textbf{Our contributions are as follows:}

\begin{itemize}
    \item We propose ATAC, a novel test-time adversarial defense method that is, to the best of our knowledge, the first method that directly corrects visual embeddings in CLIP's feature space.
    \item We demonstrate that ATAC achieves state-of-the-art robustness across 13 classification benchmarks, outperforming both adversarial fine-tuning methods and test-time defenses by nearly 50\% on average, with minimal computational overhead.
    \item Further analysis of the effectiveness of ATAC reveals a fundamental flaw of untargeted gradient-based attacks that our method can exploit; however, we show that ATAC still achieves state-of-the-art robustness when this flaw is eliminated.
    \item Finally, we show that ATAC achieves nontrivial robustness even against adaptive attacks, surpassing other test-time defenses.
\end{itemize}

%Under standard (early-stopped) evaluation, it achieves state-of-the-art robustness while preserving high clean accuracy.
% Further analysis under non–early-stopped attacks reveals that ATAC can even outperform clean-sample baselines—serving not as a main selling point but as empirical evidence of ATAC’s ability to semantically realign adversarial embeddings. Together, these results highlight a new paradigm for test-time adversarial defense: \emph{direct semantic correction in CLIP’s embedding space}.
% \paragraph{Feature-level correction paradigm.} We introduce the first test-time adversarial defense that directly corrects image embeddings in CLIP’s feature space, rather than modifying pixels (TTC) or prompts (R-TPT).

% \paragraph{Augmentation-driven semantic recovery.} We leverage augmentation-induced drift vectors and a cosine-consistency gate to infer a robust semantic recovery direction, overcoming the limitations of naive test-time ensembling.

% \paragraph{Empirical validation and insight.} We demonstrate state-of-the-art robustness across 13 benchmarks and provide a principled analysis showing that the observed “robust > clean” cases under specific settings substantiate ATAC’s semantic alignment capability.
\section{Related Work}

\paragraph{Adversarial robustness in VLMs.}
While adversarial fragility is well documented in standard deep neural networks~\cite{ilyas2019adversarialexamplesbugsfeatures,goodfellow2015explainingharnessingadversarialexamples,carlini2017evaluatingrobustnessneuralnetworks}, recent work has increasingly focused on vision-language models (VLMs) such as CLIP~\cite{ma2025safetyscalecomprehensivesurvey,zhou2024revisiting,mao2023understanding}. For VLMs, defenses can be organized into three (often overlapping) families. (i) \emph{Fine-tuning–based robustness} adversarially fine-tunes the image/text encoders or the contrastive objective to enlarge margins~\cite{schlarmann2024robustclip,mao2023understanding,wang2024pretrainedmodelguidedfinetuning}, but this is costly at the scale of foundation models and can erode zero-shot transfer~\cite{xing2025clipstrongfightback}. (ii) \emph{Prompt-side adaptation} adjusts text and/or visual prompts; this includes \emph{training-time} prompt tuning (e.g., soft prompts) and \emph{test-time} prompt tuning (TPT) that adapts prompts for each input without retraining~\cite{Zhou2021LearningTP,tong2024testtimealignment,wang2024tapttesttimeadversarialprompt,li2024promptwordboostadversarial}. In particular, TAPT~\cite{wang2024tapttesttimeadversarialprompt} and R-TPT~\cite{sheng2025rtptimprovingadversarialrobustness} sit at the intersection of prompt methods and test-time adaptation. (iii) \emph{Test-time defenses beyond prompts} avoid parameter updates by acting on the inference pipeline by applying, for example, stochastic transformations and ensembling to stabilize decisions, or image-side counterattacks that optimize the input at inference to push predictions back to the ground-truth class~\cite{pérez2021enhancingadversarialrobustnesstesttime,xing2025clipstrongfightback,cohen2019certifiedadversarialrobustnessrandomized}. 

\paragraph{Test-time Defenses for VLMs.}
Vision-language models (VLMs), such as CLIP, inherit strong zero-shot capability and vulnerability to adversarial perturbations in their joint embedding space.
Although conventional test-time adaptation (TTA) for VLMs has been extensively studied~\cite{shu2022testtimeprompttuningzeroshot,feng2023diversedataaugmentationdiffusions,hassan2024alignpromptstesttimeprompting}, defending against adversarial attacks at test time remains a relatively new direction.
Recent test-time defenses can be broadly categorized into three fronts: \emph{image-side}, \emph{prompt-side}, and \emph{latent-side} approaches.

On the \textbf{image side}, TTC~\cite{xing2025clipstrongfightback} formulates a PGD-style counter-attack at inference to escape the ``toxic'' adversarial basin and recover semantics, achieving solid robustness, albeit with nontrivial computational overhead.
Deng~\emph{et al.}~\cite{deng2025fptnoisedynamicsceneawarecounterattack} proposed FPT-Noise, a dynamic scene-aware test-time defense that adaptively injects counterattack noise guided by a feature perception threshold and regulates perturbation strength via scene-aware control, followed by test-time ensembling to suppress residual noise.
Liu~\emph{et al.}~\cite{liu2025self} further introduced Self-Calibrated Consistency (SCC), enforcing semantic and spatial consistency across augmented views to correct adversarially perturbed embeddings, substantially improving CLIP’s zero-shot robustness without retraining.

On the \textbf{prompt side}, APT~\cite{li2024promptwordboostadversarial} introduces prompt tuning into adversarial defense, but requires optimization through backpropagation.
TAPT~\cite{wang2024tapttesttimeadversarialprompt} learns bimodal (visual and textual) defensive prompts for each test sample through multi-view entropy minimization and distribution alignment, while R-TPT~\cite{sheng2025rtptimprovingadversarialrobustness} reformulates the classical objective to further enhance robustness.
Beyond adversarial robustness, more extensive test-time adaptations or parameter-efficient fine-tuning for CLIP (e.g., TPT and prompt learning variants~\cite{shu2022testtimeprompttuningzeroshot,feng2023diversedataaugmentationdiffusions,hassan2024alignpromptstesttimeprompting}, CLIP-Adapter~\cite{gao2021clip}) demonstrate that lightweight test-time interventions—either on features or prompts—can substantially alter VLM behavior without retraining.

On the \textbf{latent side}, CLIPure~\cite{zhangclipure} follows a purification-based approach, denoising CLIP embeddings directly in the latent space. 
Building upon a stochastic differential equation (SDE) framework that bridges attack and purification, it models latent likelihoods via diffusion priors (CLIPure-Diff) or cosine similarity (CLIPure-Cos). 

In this work, we introduce a novel latent-side test-time defense that, instead of purification, optimizing pixels, or tuning prompts, performs semantic correction by leveraging augmentation-induced drift vectors to realign adversarial embeddings toward their true semantics.

\section{Preliminaries and Notation}

\paragraph{Zero-shot classification with CLIP.} CLIP is a vision-language model comprising two modules: an image encoder $E_I$ and a text encoder $E_T$. These modules have been trained on 400M image-text pairs to align images to their corresponding texts via cosine similarity.

Given a $k$-class image classification problem with labels $y_1, \ldots, y_k$, text prompts $\{{T(y_i)\}}_{i=1}^k$ are constructed that represent each class, e.g., ``\texttt{A photo of a \{label\}.}'' Let $t_i = E_T(T(y_i))$, denote the encoded representation of the text prompts, and $f_x = E_I(x)$ denote the encoded representation of an
image $x$.
Then, CLIP predicts the label that maximizes the cosine similarity between its embedding and the embedding of the image:
\begin{equation}
\argmax_{i}
\frac{
\exp(\cos(f_x, t_i)/T)
}{
\sum_{j=1}^{k} \exp(\cos(f_x, t_j)/T)
},
\label{eq:clip-classification}
\end{equation}
where $\cos(\cdot)$ represents the cosine similarity operation and $T$ is a temperature parameter, typically set to 0.01.

\paragraph{Adversarial attacks on CLIP.} CLIP is highly vulnerable to adversarial attacks~\cite{mao2023understanding}. The goal of the attacker is to find a small perturbation $\delta$, such that $\delta$ is $\epsilon$-bounded in the $L_p$-ball, i.e., $\lVert\delta\rVert_p \leq \epsilon$, and the image $x + \delta$ is misclassified by CLIP.\@ For the sake of simplicity, we omit the projection of $x + \delta$ to the pixel space in all notations.

In a white-box setting, the attacker has access to the model, its gradients, and the ground truth label $y_c$ of the image $x$. To find an adversarial perturbation, the attacker solves the optimization problem
\begin{equation}
\delta^* =
\argmax_{\delta \in \mathcal{S}}
\mathcal{L}(x+\delta, y_c),
\label{eq:white-box-attack}
\end{equation}
where $S=\{\delta : \lVert \delta \rVert_p \leq \epsilon \}$, $\mathcal{L}(\cdot)$ measures the classification loss of the model, and the adversarial input is $x_a = x + \delta^*$.

\cref{eq:white-box-attack} can be indirectly approximated by the Projected Gradient Descent (PGD)~\cite{madry2019deeplearningmodelsresistant} algorithm:
\begin{equation}
x^{t} = \prod_{x + S}(x^{t-1} + \gamma\text{sgn}(\nabla_{x}\mathcal{L}(x, y_c))) \quad (t=1\ldots T),
\label{eq:pgd}
\end{equation}
where $T$ is the number of attack steps, $\gamma$ is the step size, $x^T$ is the adversarial example, and $x_0=x$, or, for PGD with random start, $x_0 = x + \delta_0$, where $\delta_0 \sim \mathcal{U}(-\epsilon, \epsilon)$.
% We'll see based on requirements and remaining space:
% PGD formula? -- added above for later reference
% Targeted attacks?

\section{The ATAC Framework}

In this section, we introduce \textbf{A}ugmentation-based \textbf{T}est-time \textbf{A}dversarial \textbf{C}orrection (ATAC), a novel test-time adversarial defense method for CLIP.\@

\subsection{Motivation}%
\label{sec:motivation}
Let us begin by highlighting three key results from related work along with their shortcomings that motivate our method.

\paragraph{Test-time counterattacks.}
Although adversarial finetuning has been shown to substantially increase the robustness of CLIP, it requires costly training procedures. To mitigate this, Xing et al.~\cite{xing2025clipstrongfightback} explore test-time counterattacks that enable CLIP to defend itself without additional training. Their counterattack, also implemented by an adversarial attack, aims to maximize the $L_2$ distance between the original and the counterattacked image in the embedding space, thereby allowing the input to escape the adversarial ``toxic region'' in the embedding space, where it was moved by the original (malicious) attack.

However, this approach has two key limitations. Firstly, during inference, it relies on computationally costly counterattacks with sensitive hyperparameters. Secondly, the objective is only aimed at maximizing embedding shift, without any emphasis on restoring the original semantics of the input. Therefore, even if the counterattack succeeds in moving the embedding away from the adversarial region, this movement may not be towards the semantically correct direction.

\paragraph{Augmentations against adversarial attacks.}
Although recent work shows that the embedding space of CLIP remains relatively stable towards common image transformations (e.g., flipping, rotations, and color jittering)~\cite{dahal2025embeddingshiftdissectionclip}, extensive studies show that adversarial attacks are highly sensitive to such augmentations~\cite{ilyas2019adversarialexamplesbugsfeatures,mao2021adversarialattacksreversiblenatural,lindqvist2021delvingpixelsadversarialsamples,zeng2019adversarialattacksimagespace}. In other words, augmentations can mitigate the effect of adversarial attacks.

This observation motivates test-time transformation ensembling (TTE)~\cite{pérez2021enhancingadversarialrobustnesstesttime}, where predictions over augmented views are aggregated to form the final prediction. However, as shown in \cref{tab:pgd_eps4}, TTE yields limited robustness gains, leading us to hypothesize that augmentations alone are not enough to mitigate the effects of adversarial attacks.

\paragraph{Robust test-time prompt tuning.}
R-TPT~\cite{sheng2025rtptimprovingadversarialrobustness} uses numerous augmentations to find views of the input image with low-entropy predictions. In order to ignore adversarial or outlier views, they ensemble these predictions using a weighting scheme based on feature-level nearest neighbors to obtain the corrected prediction.

However, only aggregating predictions may allow low-entropy incorrect predictions of adversarial views to still mislead the method. Moreover, their weighting scheme is not an explicit way of recovering the semantics of the original input, as feature representations of incorrectly classified views can form tight clusters.

\subsection{Our Method}
\label{sec:atac_intro}

To address the limitations of previous work, we propose \textbf{A}ugmentation-based \textbf{T}est-time \textbf{A}dversarial \textbf{C}orrection (ATAC). Our method uses the latent representations of augmented views to explicitly estimate a semantic recovery direction for adversarial inputs, and a cosine-consistency gate to control the correction process and avoid the over-correction of clean samples. \cref{fig:atac_overview} shows an overview of our method.

Formally, given an input image $x$, we first apply $n$ different transformations that yield the augmented views $x_1, \ldots, x_n$, and, using the image encoder of CLIP, get their encoded representations $f_{x_1}, \ldots, f_{x_n}$ along with the encoded representation of the original input $f_x$. We then compute the latent drifts $d_1, \ldots, d_n$ and the mean drift $\bar{d}$ as
\begin{equation}
\{{d_i = f_x - f_{x_i}\}}_{i=1}^n,\quad \bar{d} = \frac{1}{n}\sum_{i=1}^{n}d_i.
\end{equation}

We assess the \textit{directional consistency} of the latent drift vectors with the mean drift as
\begin{equation}
\tau = \frac{1}{n}\sum_{i=1}^{n}\cos(d_i, \bar{d}),
\end{equation}
where $\cos$ is the cosine similarity operation. Lower $\tau$ values indicate scattered drift vectors, while higher values indicate that drift vectors point towards the same direction in the latent space. Since augmentations can mitigate the effect of adversarial attacks (as discussed in \cref{sec:motivation}), we expect adversarial inputs to yield directionally consistent drift vectors. In this case, $\bar{d}$ represents a \textit{semantic recovery direction} that we use to recover the original semantics of the attacked image.

On the other hand, we expect that clean inputs yield low $\tau$ values. To avoid modifying the latent representations of clean images, we introduce a gating threshold $\tau^*$, and perform the test-time correction 
%$f^* = f_x - \alpha \bar{d}$ if $\tau > \tau^*$, 
%
\begin{equation}
f^* = f_x + \alpha \bar{d} \quad\text{if}\quad \tau > \tau^*,
\end{equation}
otherwise we keep the original image embedding, i.e., $f^* = f_x$. We then normalize\footnote{Note that, when $f^* = f_x$ (i.e., $f_x$ is not corrected), this normalization has no effect, since $f_x$ is already normalized.} $f^*$ and treat it as the visual embedding for downstream tasks, such as classification.% via \cref{eq:clip-classification}.

Our method builds upon the robust embedding space of CLIP and the ability of augmentations to mitigate adversarial attacks. We address the limitations of related works by explicitly estimating a semantic recovery direction in CLIP's embedding space, using the representations of augmented views. ATAC is training-free and only requires $n$ forward passes at inference, making it highly efficient compared to other test-time defenses, as shown in \cref{tab:inference_time}.

\begin{table}[t]
\centering
\begin{tabular}{lr@{{\color{gray}$\pm$}}l}
Method & \multicolumn{2}{c}{Inference time (s)} \\
\midrule
CLIP (no defense)~\cite{Radford2021LearningTV} & 3.63 & \color{gray}1.67 \\
%TTE~\cite{pérez2021enhancingadversarialrobustnesstesttime} & ... & \color{gray}... \\
TTC~\cite{xing2025clipstrongfightback}& 20.67 & \color{gray}1.96  \\
R-TPT~\cite{sheng2025rtptimprovingadversarialrobustness}& 916.33 & \color{gray}111.06  \\
\textbf{ATAC (ours)} & 18.41 & \color{gray}0.40  \\
\end{tabular}
\caption{Inference times of the original CLIP and test-time defense methods on 1000 images, using a single RTX A6000 GPU. Results are averaged over 6 datasets, with standard deviations indicated.}
\label{tab:inference_time}
\end{table}

\begin{table*}[htbp]

\centering
\scriptsize
\setlength{\tabcolsep}{5pt}

\renewcommand{\arraystretch}{1.5}

\resizebox{\textwidth}{!}{%
\begin{tabular}{
  c                                   % Dataset
  c |                                  % Rob./Acc.
  S[table-format=2.2(2)] |             % CLIP (numbers)
  *{4}{S[table-format=2.2(2)]} ||       % Adversarial Finetuning
  *{4}{S[table-format=2.2(2)]} |       % Test-time Defence
  S[table-format=+2.2, retain-explicit-plus]                % Delta (signed)
}
\toprule
\multicolumn{2}{c}{(\%)} &
\multicolumn{1}{|c|}{\multirow{2}{*}{CLIP}} &
\multicolumn{4}{c||}{\textbf{Adversarial Finetuning}} &
\multicolumn{4}{c|}{\textbf{Test-time Defense}} &
\multicolumn{1}{c}{\multirow{2}{*}{$\Delta$}} \\
% \cmidrule(lr){4-10} \cmidrule(lr){11-15}
& & &
\multicolumn{1}{c}{CLIP-FT} &
% \multicolumn{1}{c}{TeCoA$^1$} &
\multicolumn{1}{c}{TeCoA} &
% \multicolumn{1}{c}{PMG-AFT$^1$} &
\multicolumn{1}{c}{PMG-AFT} &
% \multicolumn{1}{c}{FARE$^1$} &
\multicolumn{1}{c||}{FARE} &
% \multicolumn{1}{c}{RN} &
\multicolumn{1}{c}{TTE} &
\multicolumn{1}{c}{TTC} &
\multicolumn{1}{c}{R-TPT} &
\multicolumn{1}{c|}{\textbf{ATAC (ours)}} &
\\
\midrule

\multirow{2}{*}{\textbf{CIFAR10}} &
Rob. &
\num{0.43}  &  \num{2.75}  &  \num{11.7}  &  \num{15.59}  &  \num{5.42}  &  \num{3.47(2.77)}   &  \num{28.51(0.36)}  &  \underline{\num{33.84(1.55)}}  &  \textbf{\num{91.80(0.09)}}  & \textcolor{red}{\num{+90.37}} \\
& Acc. &
\num{85.12}  &  \textbf{\num{84.90}}  &  \num{65.15}  &  \num{71.45}  &  \num{78.46}  &  \num{84.74(0.40)}   &  \num{81.18(0.07)}  &  \num{82.19(1.03)}  &  \num{79.39}  & \textcolor{blue}{\num{-5.63}} \\
\midrule

\multirow{2}{*}{\textbf{CIFAR100}} &
Rob. &
\num{0.05}  &  \num{0.67}  &  \num{9.25}  &  \num{10.80}  &  \num{4.54}  &  \num{1.37(0.96)}  &   \num{9.06(0.11)}  &  \underline{\num{18.52(1.08)}}  &  \textbf{\num{86.27(0.42)}}  & \textcolor{red}{\num{+86.22}} \\
& Acc. &
\num{57.14}  &  \textbf{\num{59.51}}  &  \num{36.30}  &  \num{41.51}  &  \num{47.38}  &  \num{58.61(0.25)}   &  \num{56.34(0.20)}  &  \num{52.69(0.94)}  &  \num{52.65}  & \textcolor{blue}{\num{-4.49}} \\
\midrule

\multirow{2}{*}{\textbf{STL10}} &
Rob. &
\num{0.16}   &   \num{3.75}   &   \num{31.83}   &   \num{35.40}   &   \num{17.59}   &   \num{32.56(11.76)}   &    \num{52.40(0.34)}   &   \underline{\num{76.33(2.46)}}   &   \textbf{\num{98.30(0.16)}}   &   \textcolor{red}{\num{+98.14}} \\
& Acc. &
\num{96.40}   &   \num{94.49}   &   \num{81.69}   &   \num{84.35}   &   \num{89.11}   &   \textbf{\num{96.26(0.04)}}   &    \num{95.83(0.03)}   &   \num{96.09(0.24)}   &   \num{93.94}   &   \textcolor{blue}{\num{-2.46}} \\
\midrule

\multirow{2}{*}{\textbf{Caltech101}} &
Rob. &
\num{0.59}   &   \num{4.81}   &   \num{21.00}   &   \num{25.03}   &   \num{10.13}   &   \num{30.19(7.92)}   &    \num{36.66(0.25)}   &   \underline{\num{68.11(0.24)}}   &   \textbf{\num{86.76(0.11)}}   &  \textcolor{red}{\num{+86.17}} \\
& Acc. &
\num{85.66}   &   \num{83.63}   &   \num{64.41}   &   \num{69.06}   &   \num{76.58}   &   \num{85.84(0.09)}   &   \num{86.15(0.08)}   &   \textbf{\num{86.62(0.62)}}   &   \num{82.00}   &  \textcolor{blue}{\num{-3.66}} \\
\midrule

\multirow{2}{*}{\textbf{Caltech256}} &
Rob. &
\num{0.12}   &   \num{1.41}   &   \num{11.76}   &   \num{13.68}   &   \num{5.09}   &   \num{23.23(7.77)}   &   \num{27.25(0.08)}   &   \underline{\num{54.45(0.71)}}   &   \textbf{\num{90.86(0.11)}}   &  \textcolor{red}{\num{+90.74}} \\
& Acc. &
\num{81.72}   &   \num{78.53}   &   \num{52.05}   &   \num{53.32}   &   \num{67.22}   &   \textbf{\num{82.48(0.08)}}   &   \num{76.59(0.12)}   &   \num{77.67(0.47)}   &   \num{79.10}   &  \textcolor{blue}{\num{-2.62}} \\
\midrule

\multirow{2}{*}{\textbf{OxfordPets}} &
Rob. &
\num{0.00}   &   \num{1.66}   &   \num{3.71}   &   \num{5.10}   &   \num{0.30}   &   \num{3.18(2.94)}    &   \num{24.64(0.53)}   &   \underline{\num{44.15(1.08)}}   &   \textbf{\num{87.46(0.19)}}   &  \textcolor{red}{\num{+87.46}} \\
& Acc. &
\num{87.44}   &   \num{84.14}   &   \num{53.94}   &   \num{56.66}   &   \num{70.10}   &   \textbf{\num{88.13(0.13)}}    &   \num{64.70(0.33)}   &   \num{84.46(0.62)}   &   \num{85.15}   &  \textcolor{blue}{\num{-2.29}} \\
\midrule

\multirow{2}{*}{\textbf{Flowers102}} &
Rob. &
\num{0.00}   &   \num{0.13}   &   \num{3.81}   &   \num{4.26}   &   \num{0.62}   &   \num{3.52(2.51)}   &   \num{13.60(0.33)}   &   \underline{\num{32.46(0.47)}}   &   \textbf{\num{85.69(0.16)}}   &  \textcolor{red}{\num{+85.69}} \\
& Acc. &
\num{65.46}   &   \num{53.37}   &   \num{27.78}   &   \num{28.88}   &   \num{41.01}   &   \num{65.20(0.23)}   &   \num{63.24(0.21)}   &   \num{62.92(0.85)}   &   \textbf{\num{64.29}}   &  \textcolor{blue}{\num{-1.17}} \\
\midrule

\multirow{2}{*}{\textbf{FGVCAircraft}} &
Rob. &
\num{0.00}   &   \num{0.00}   &   \num{0.12}   &   \num{0.06}   &   \num{0.03}   &   \num{0.43(0.43)}    &   \num{6.40(0.38)}   &   \underline{\num{7.20(0.62)}}   &   \textbf{\num{50.02(0.31)}}   &  \textcolor{red}{\num{+50.02}} \\
& Acc. &
\num{20.10}   &   \num{14.04}   &   \num{3.51}   &   \num{3.24}   &   \num{7.77}   &   \textbf{\num{20.18(0.35)}}    &   \num{15.99(0.04)}   &   \num{19.14(0.62)}   &   \num{19.65}   &  \textcolor{blue}{\num{-0.45}} \\
\midrule

\multirow{2}{*}{\textbf{StanfordCars}} &
Rob. &
\num{0.00}   &   \num{0.00}   &   \num{0.41}   &   \num{0.40}   &   \num{0.04}   &   \num{1.46(1.21)}   &   \num{12.84(0.20)}   &   \underline{\num{20.76(1.78)}}   &   \textbf{\num{70.80(0.08)}}   &  \textcolor{red}{\num{+70.80}} \\
& Acc. &
\num{52.02}   &   \num{42.11}   &   \num{15.18}   &   \num{16.79}   &   \num{32.09}   &   \textbf{\num{52.73(0.31)}}    &   \num{41.52(0.15)}   &   \num{61.75(0.24)}   &   \num{51.41}   &  \textcolor{blue}{\num{-0.61}} \\
\midrule

\multirow{2}{*}{\textbf{Country211}} &
Rob. &
\num{0.00}   &   \num{0.00}   &   \num{0.19}   &   \num{0.24}   &   \num{0.02}   &   \num{0.24(0.15)}    &   \underline{\num{2.44(0.15)}}   &   \num{0.42(0.24)}   &   \textbf{\num{65.55(0.25)}}   &  \textcolor{red}{\num{+65.55}} \\
& Acc. &
\num{15.25}   &   \num{12.07}   &   \num{3.66}   &   \num{3.34}   &   \num{6.58}   &   \num{14.66(0.14)}    &   \num{11.99(0.01)}   &   \num{13.40(0.62)}   &   \textbf{\num{16.45}}   &  \textcolor{red}{\num{+1.20}} \\
\midrule

\multirow{2}{*}{\textbf{Food101}} &
Rob. &
\num{0.00}   &   \num{0.04}   &   \num{1.35}   &   \num{2.12}   &   \num{0.24}   &   \num{5.31(4.09)}   &   \num{17.89(0.13)}   &   \underline{\num{39.97(1.25)}}   &   \textbf{\num{96.11(0.07)}}   &  \textcolor{red}{\num{+96.11}} \\
& Acc. &
\num{83.88}   &   \num{64.86}   &   \num{21.90}   &   \num{27.97}   &   \num{41.98}   &   \textbf{\num{83.96(0.01)}}    &   \num{80.00(0.07)}   &   \num{83.41(0.47)}   &   \num{82.87}   &  \textcolor{blue}{\num{-1.01}} \\
\midrule

\multirow{2}{*}{\textbf{EuroSAT}} &
Rob. &
\num{0.00}   &   \num{0.00}   &   \num{10.71}   &   \num{10.36}   &   \num{7.34}   &   \num{0.11(0.09)}    &   \underline{\num{13.57(0.12)}}   &   \num{6.46(1.03)}   &   \textbf{\num{66.57(0.06)}}   &  \textcolor{red}{\num{+66.57}} \\
& Acc. &
\num{42.59}   &   \num{27.64}   &   \num{17.53}   &   \num{19.19}   &   \num{18.22}   &   \num{44.38(1.62)}   &   \textbf{\num{53.24(0.09)}}   &   \num{21.83(1.22)}   &   \num{37.28}   &  \textcolor{blue}{\num{-5.31}} \\
\midrule

\multirow{2}{*}{\textbf{DTD}} &
Rob. &
\num{0.11}   &   \num{0.00}   &   \num{5.16}   &   \num{5.21}   &   \num{2.50}   &   \num{7.16(2.32)}   &   \num{11.40(0.28)}   &   \underline{\num{26.97(1.03)}}   &   \textbf{\num{76.06(0.58)}}   &  \textcolor{red}{\num{+75.95}} \\
& Acc. &
\num{40.64}   &   \num{36.49}   &   \num{20.11}   &   \num{17.29}   &   \num{28.03}   &   \textbf{\num{41.35(0.29)}}   &   \num{35.69(0.08)}   &   \num{42.66(0.41)}   &   \num{38.46}   &  \textcolor{blue}{\num{-2.18}} \\
\midrule
\midrule

\multirow{2}{*}{Avg.} &
Rob. &
\num{0.11}   &   \num{1.17}   &   \num{8.54}   &   \num{9.87}   &   \num{4.14}   &   \num{8.63(3.23)}   &   \num{19.74(0.05)}   &   \underline{\num{33.05(0.47)}}   &   \textbf{\num{80.94(0.02)}}   &  \textcolor{red}{\num{+80.83}} \\
& Acc. &
\num{62.57}   &   \num{56.60}   &   \num{35.63}   &   \num{37.93}   &   \num{46.50}   &   \textbf{\num{62.96(0.13)}}   &   \num{58.65(0.06)}   &   \num{60.37(0.36)}   &   \num{60.20}   &  \textcolor{blue}{\num{-2.37}} \\

\bottomrule
\end{tabular}
}% end resizebox
%\vspace{5pt}
\caption{Classification accuracy (\%) on both adversarial images (Rob.) under a 10-step PGD attack with $\epsilon=4/255$ and clean images (Acc.) across 13 datasets. We report the mean and standard deviation for test-time methods over 3 runs. The best robust and clean accuracies among adversarial defenses are indicated in bold, with the second best robust accuracies underlined for ease of comparison. The last column reports the gains of our method w.r.t. original CLIP without any finetuning or test-time operations.}
% \caption{Classification accuracy (\%) on both adversarial images (Rob.) under 10-step PGD attack at $\epsilon_a = 4/255$ and clean images (Acc.) across 13 datasets. Weights and gradients of the deployed model are assumed to be known to the threat model. Comparison is made among our paradigm and other test-time defences adapted from existing adversarial studies, test-time counter attack and test-time prompt tuning, with finetuning-based models implemented as a reference. The superscripts of the model names indicate the attack budget used for generating adversarial images in the phase of adversarial finetuning. We report the mean and standard deviation for test-time methods over 3 runs. The last column reports the gains w.r.t. original CLIP without any finetuning or test-time operations.}
\label{tab:pgd_eps4}
\end{table*}

\begin{figure*}[htbp]
    \centering
    \includegraphics[width=1\linewidth]{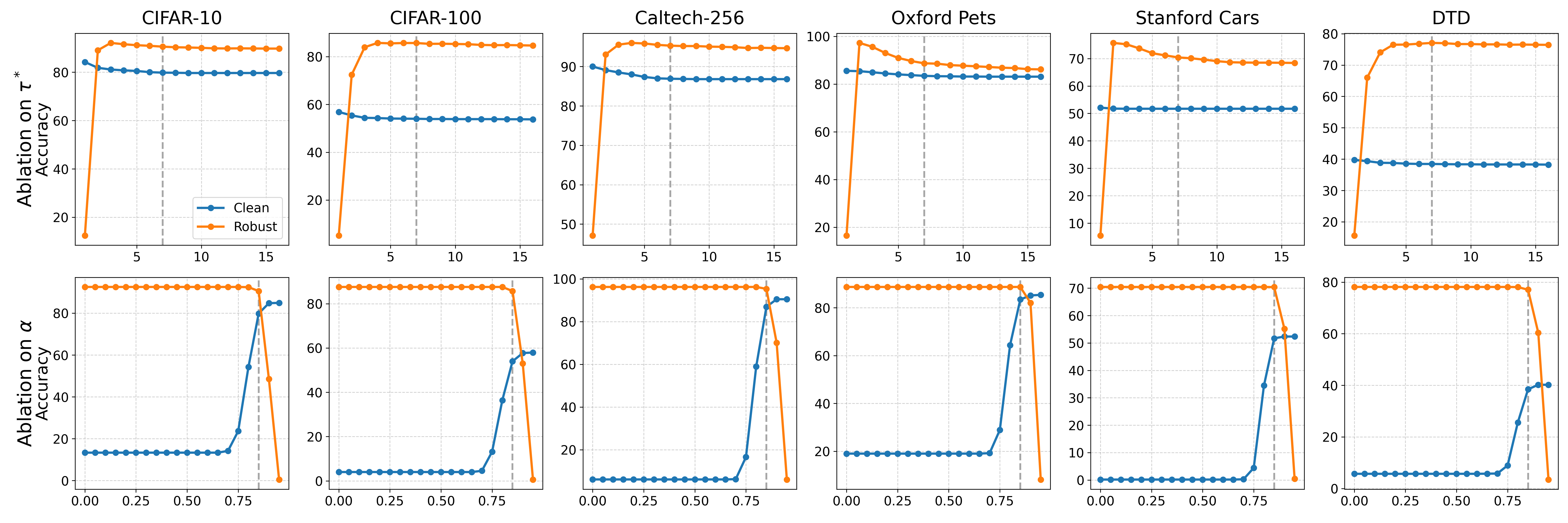}
    \caption{Ablations on the cosine-consistency threshold $\tau^*$ (top row) and the correction step size $\alpha$ (bottom row) on 6 datasets. \blue{Blue} lines show accuracy on clean images and \textcolor{orange}{orange} lines represent accuracy on adversarial images in each parameter setting. Vertical \textcolor{gray}{gray} lines represent the parameter settings adopted in our experiments.
    % Clean and robust accuracies of ATAC across different hyperparameter settings.The \textcolor{blue}{blue} lines represent accuracy on clean images and the \textcolor{orange}{orange} lines represent accuracy on adversarial images. The vertical \textcolor{gray}{gray} lines represent the actual parameter settings adopted in main text. Each experiment is conducted on 2,000 samples drawn from each dataset.
    } 
    \label{fig:ablations}
\end{figure*}

\section{Results}
\label{sec:results}

\subsection{Experimental Setup}
\label{sec:expsetup}
\paragraph{Datasets.}
We closely follow the experimental setup of~\cite{xing2025clipstrongfightback} and conduct our experiments on 13 datasets, which include general object recognition datasets CIFAR10~\cite{Krizhevsky2009LearningML}, CIFAR100~\cite{Krizhevsky2009LearningML}, STL10~\cite{Coates2011AnAO}, Caltech101~\cite{FeiFei2006OneshotLO} and Caltech256~\cite{Griffin2007Caltech256OC}, fine-grained recognition datasets OxfordPets~\cite{parkhi12a}, Flowers102~\cite{Nilsback2008AutomatedFC}, Food101~\cite{Bossard2014Food101M}, StanfordCars~\cite{Krause20133DOR}, the scene recognition dataset Country211~\cite{Radford2021LearningTV}, and domain-specific datasets FGVCAircraft~\cite{Maji2013FineGrainedVC}, EuroSAT~\cite{Helber2017EuroSATAN}, DTD~\cite{Cimpoi2013DescribingTI}. Similar to~\cite{xing2025clipstrongfightback}, we used the pre-processing pipeline of CLIP~\cite{Radford2021LearningTV} for all datasets.

\paragraph{Implementation Details.}
Following~\cite{xing2025clipstrongfightback}, we used the official pre-trained CLIP ViT-B/32~\cite{Radford2021LearningTV} in our implementation.
In all experiments, we used the cosine-consistency threshold $\tau^* = 0.85$ and the correction step size $\alpha=7$. We used $n=5$ augmentations in our framework, namely horizontal flip and rotations with degrees $\pm15^\circ$ and $\pm30^\circ$. We present ablations over these parameters in \cref{sec:ablations} and the Appendix (\cref{sec:appx_ablations}).

\paragraph{Adversaries.}
We present our results against the PGD adversary (\cref{eq:pgd}) under the $L_\infty$ norm. Unless otherwise specified, we used $\epsilon=4/255$, $T=10$, $\gamma=1/255$ and random start in all our attacks. We present results against further adversaries, such as the Carlini-Wagner attack~\cite{carlini2017evaluatingrobustnessneuralnetworks} and AutoAttack~\cite{croce2020reliableevaluationadversarialrobustness}, in the Appendix (\cref{sec:appx_other_attacks}), with similarly excellent results.

\paragraph{Baselines.}
We compare ATAC with several test-time defenses for CLIP, as well as adversarial fine-tuning methods. Among test-time defenses, we use Test-time Transformation Ensembling (TTE)~\cite{pérez2021enhancingadversarialrobustnesstesttime} with 9 augmentations (horizontal flip, 4 crops, and horizontal flip with 4 crops), Test-time Counterattacks (TTC)~\cite{xing2025clipstrongfightback} with a PGD-style counterattack of 5 steps, $\epsilon_{ttc}=4/255$, $\tau^*_{ttc} = 0.2$, and $\beta=2$, as well as Robust Test-time Prompt Tuning (R-TPT)~\cite{sheng2025rtptimprovingadversarialrobustness} with 64 augmentations, and a 1-step Adam optimizer with a learning rate of 0.005. 

% To compare ATAC with adversarial fine-tuning methods, we use variants of TeCoA~\cite{mao2023understanding}, PMG-AFT~\cite{wang2024pretrainedmodelguidedfinetuning}, FARE~\cite{schlarmann2024robustclip}, and the regular CLIP image encoder (CLIP-FT) that were adversarially fine-tuned using the Tiny ImageNet dataset~\cite{le2015tiny} by~\cite{xing2025clipstrongfightback} with an attack budget of $\epsilon=4/255$.
To compare ATAC with adversarial fine-tuning methods, we use the baselines of~\cite{xing2025clipstrongfightback}: TeCoA~\cite{mao2023understanding}, PMG-AFT~\cite{wang2024pretrainedmodelguidedfinetuning}, and FARE~\cite{schlarmann2024robustclip} that were adversarially fine-tuned using the Tiny ImageNet dataset~\cite{le2015tiny} with an attack budget of $\epsilon=4/255$, as well as the regular CLIP image encoder (CLIP-FT) fine-tuned on Tiny ImageNet without adversarial training.

\subsection{ATAC for Adversarial Robustness}
\label{sec:clip_results}
We evaluate all defense methods across 13 datasets against the PGD attack and present the results in~\cref{tab:pgd_eps4}. For most datasets, Tiny ImageNet-based adversarial fine-tuning yields minimal robustness gains, showing a significant limitation of this approach and the superiority of test-time defenses. 

Among test-time defenses, \emph{ATAC clearly stands out} in terms of robustness by consistently achieving the highest robust accuracy across all methods. ATAC significantly improves the robustness of undefended CLIP up to 90\% in some cases, without any fine-tuning or costly test-time optimization. Furthermore, in all cases, ATAC substantially outperforms the robustness of previous state-of-the-art test-time defenses such as R-TPT and TTC by as much as 70\%, and nearly 50\% on average.

As expected, adversarial defense methods usually incur a penalty when it comes to their performance on clean samples. This phenomenon, known as the robustness-accuracy trade-off, is also present with ATAC: in order to gain robustness, our method suffers a minor loss in accuracy as a result of correcting some clean examples. For ATAC, the decrease in clean accuracy is in line with previous methods. However, the increase in robustness is outstanding, and the resulting robust accuracy sometimes eclipses clean performance. We further investigate this phenomenon in \cref{sec:main_robust_surpass_clean}.

\subsection{Ablation Study}
\label{sec:ablations}
In this section, we investigate the effect of the cosine-consistency threshold $\tau^*$ and the correction step size $\alpha$ separately, while keeping the other parameters of ATAC unchanged. We conduct ablations on six datasets and follow the experimental setup described in~\cref{sec:expsetup}. Each parameter setting is evaluated on 2000 samples drawn from each dataset. We present our results in~\cref{fig:ablations}, along with further ablations on the augmentations used in ATAC in the Appendix (\cref{sec:appx_ablations}).

The effect of $\alpha$ is minimal across all datasets, particularly when it is sufficiently large ($\geq 3$). This is likely a result of the normalization of the corrected visual embedding. This result further demonstrates that the effectiveness of ATAC is based primarily on the correct estimation of the semantic recovery direction and is insensitive to the value of $\alpha$.

However, the value of $\tau^*$ is crucial. When $\tau^*$ is low (i.e., $\leq 0.7$), clean samples are not protected from the correction mechanism, leading to extremely low clean accuracies. As $\tau^*$ increases, fewer clean samples are corrected unnecessarily, but the ratio of corrected adversarial samples also decreases, leading to an increase in clean accuracy and a decrease in robust accuracy. We found $\tau^*=0.85$ to yield the best balance between robustness and accuracy across all datasets, which is further supported by our analysis of augmentations and $\tau$ distributions in the Appendix (\cref{sec:appx_tau_dist}).

%The earlier section shows that adversarial accuracy surpasses clean accuracy in almost all cases, in fact, the trade-off now goes the other way (higher robustness, lower accuracy). We investigate this phenomenon here, the current setup is good, include the following settings:
%\begin{itemize}
%    \item $\epsilon = 127.5/255$ or $\epsilon = 255/255$, i.e., very large %$\epsilon$
%    \item Early stopped attack.
%    \item Unsupervised attack.
%    \item Targeted attack. (?)
%    \item Transfer attack. (?)
%\end{itemize}

%We should visualize these settings in an easy-to-follow plot as well. \red{Around 1-1.5 pages.}

% \input{tables/extreme}
\begin{table*}[t]
\centering
\setlength{\tabcolsep}{7pt}
\renewcommand{\arraystretch}{1.4}
\resizebox{\textwidth}{!}{%
% \begin{tabular}{l cc cc cc cc cc cc}
\begin{tabular}{l || cc || cc | cc | cc | cc}
\toprule
\multirow{2}{*}{\textbf{Dataset}} & 
% \multicolumn{2}{c}{Clean acc.} &
\multicolumn{2}{c||}{PGD $\epsilon=\sfrac{4}{255}$} &
\multicolumn{2}{c|}{PGD $\epsilon=\sfrac{127.5}{255}$} &
\multicolumn{2}{c|}{Early-stopped PGD} &
\multicolumn{2}{c|}{Unsupervised PGD} &
\multicolumn{2}{c}{Targeted PGD} \\
& 
% CLIP & ATAC &
CLIP & ATAC &
CLIP & ATAC &
CLIP & ATAC &
CLIP & ATAC &
CLIP & ATAC \\
\midrule
\textbf{CIFAR-10} & 
% 85.22 & 79.25 &
0.00 & 91.80 &
0.00 (\red{+0.00}) & 98.25 (\red{+6.45}) &
0.00 (\red{+0.00}) & 59.70 (\blue{-32.10}) &
17.48 (\red{+17.48}) & 54.95 (\blue{-36.85}) &
0.05 (\red{+0.05}) & 53.50 (\blue{-38.30})
\\
\midrule
\textbf{CIFAR-100} &
0.00 & 86.27 &
0.00 (\red{+0.00}) & 95.05 (\red{+8.78}) &
0.00 (\red{+0.00}) & 44.52 (\blue{-41.75}) &
5.55 (\red{+5.55}) & 21.52 (\blue{-64.75}) &
0.07 (\red{+0.07}) & 18.02 (\blue{-68.25}) 
\\
\midrule
\textbf{STL-10} &
0.05 & 98.30 &
0.00 (\blue{-0.05}) & 98.72 (\red{+0.42}) &
0.05 (\red{+0.00}) & 92.25 (\blue{-6.05}) &
25.10 (\red{+25.05}) & 84.88 (\blue{-13.42}) &
0.47 (\red{+0.42}) & 88.20 (\blue{-10.10})
\\
\midrule
\textbf{Flowers102} &
0.00 & 85.69 & 
0.00 (\red{+0.00}) & 92.15 (\red{+6.46}) &
0.00 (\red{+0.00}) & 59.62 (\blue{-26.07}) &
3.45 (\red{+3.45}) & 30.07 (\blue{-55.62}) &
0.25 (\red{+0.25}) & 24.70 (\blue{-60.99}) 
\\
\midrule
\textbf{FGVAircraft} &
0.00 & 50.02 &
0.00 (\red{+0.00}) & 68.08 (\red{+18.73}) &
0.00 (\red{+0.00}) & 18.96 (\blue{-30.39}) &
0.63 (\red{+0.63}) & 10.56 (\blue{-38.79}) &
0.00 (\red{+0.00}) & 8.28 (\blue{-41.74})
\\
\midrule
\textbf{DTD} &
0.11 & 76.06 &
0.00 (\blue{-0.11}) & 80.16 (\red{+3.09}) &
0.11 (\red{+0.00}) & 49.73 (\blue{-27.34}) &
8.03 (\red{+7.92}) & 23.09 (\blue{-53.98}) &
1.06 (\red{+0.95}) & 19.95 (\blue{-56.11})
\\
\midrule
\midrule
\textbf{Avg.} &
0.03 & 81.36 &
0.00 (\blue{-0.03}) & 88.74 (\red{+8.45}) &
0.03 (\red{+0.00}) & 54.13 (\blue{-26.16}) &
10.04 (\red{+10.01}) & 37.51 (\blue{-42.78}) &
0.21 (\red{+0.18}) & 35.44 (\blue{-45.92})
\\
\bottomrule
\end{tabular}
}
\caption{Robust accuracies of undefended CLIP and ATAC under specially designed attack scenarios. Values in parentheses indicate the change compared to the robust accuracies of the original methods against the untargeted $\epsilon=4/255$ PGD attack, with values in \red{red} and \blue{blue} representing increased and decreased robustness values, respectively. %Each experiment is conducted on 4,000 samples drawn from each dataset.
% Performance of No Defense vs. ATAC under specially designed attack settings. "es" denotes early stopping (budget $\epsilon_\alpha=4/255$}). Values in parentheses show the change relative to the corresponding PGD eps=4/255 result — \red{red} for increase, \blue{blue} for decrease, black for no change. All results are averaged over 4000 samples per dataset.
}
\label{tab:extreme}
\end{table*}

\section{On the Robustness-Accuracy Trade-off}
\label{sec:main_robust_surpass_clean}

Our main results in~\cref{sec:results} show an unusual phenomenon: the robust accuracy of ATAC often exceeds not only its clean accuracy, but also that of CLIP. This is generally considered impossible in the adversarial robustness literature. In this section, we explore a possible source of this phenomenon and how it relates to the effectiveness of our method.

% We hypothesize that the root cause of this phenomenon is a weakness of the 
We hypothesize that this phenomenon stems from a weakness of the attack objective, namely that the attack relies too much on the ground truth label. As gradient-based, untargeted attacks maximize the loss of the true label, they move the embedding of the input away from the correct decision region along the ``path of least resistance''. However, the embedding shift introduced by the attack carries hidden directional information pointing away from the true class.

Our method is able to exploit this hidden label-dependent directional information by estimating its reverse using augmentations. Continuing our hypothesis, if one were to reduce the amount of label-dependent information the attacker could use, estimating a semantic recovery direction would be more difficult, leading to a degradation in the performance of ATAC. 

\subsection{Experimental Setup}
In order to test whether ATAC \emph{can} and \emph{does} exploit the label-dependent directional information introduced by the attacks, we design four extreme settings.

\paragraph{Increasing Label-Dependent Information.} We use PGD with $\epsilon = 127.5/255$ to maximize the amount of label-dependent information introduced by the attack. We design this setting to test the reverse of our hypothesis, i.e., introducing more label-dependent information leads to a more consistent estimation of the semantic recovery direction.

\paragraph{Early Stopping.} We use PGD with early stopping, i.e., we stop the optimization of~\cref{eq:pgd} at the earliest $t$ where $x^t$ is misclassified. This reduces the aforementioned shift in the embedding space, making the drift vectors more scattered, leading to inconsistent estimates of the recovery direction.

\paragraph{Unsupervised Attack.} We use a PGD attack that aims to maximize the $L_2$-distance between the visual features of the original and the attacked images. Although this attack introduces large shifts and thereby more consistent drift vectors, the estimated recovery direction is not guaranteed to recover the original semantics due to the attack objective completely omitting label supervision.

\paragraph{Targeted Attack.} We use a targeted PGD attack that aims to create a perturbation $\delta$ that minimizes $\mathcal{L}(x+\delta, y_t)$ for a target label $y_t \neq y_c$. Due to not having access to the true label, this attack cannot exploit the ``path of least resistance'' in the embedding space, creating smaller embedding shifts similar to the early-stopped attack.

\subsection{Results}

\cref{tab:extreme} shows the robustness of CLIP and ATAC in the four extreme settings across six datasets. We used 4000 samples from the test set of each dataset for all evaluations. The results clearly show that, as the true label-based supervision is limited, the attacks become harder to correct for our method, resulting in lower robust accuracies. In fact, the phenomenon that robust accuracy eclipses clean performance completely disappears. In contrast, when the attack budget $\epsilon$ is large, and therefore the influence of the true label is increased, the semantic recovery direction becomes easier to estimate, and ATAC achieves even higher robust accuracies. This confirms our hypothesis that unsupervised, gradient-based attacks introduce easy-to-estimate shifts in the embedding space by relying too much on the ground truth label. These results also show that ATAC can and indeed does exploit this hidden directional information.

On the other hand, even in scenarios where the attack has limited or no access to the true labels, ATAC still achieves robust accuracies that are comparable and in most cases superior to those of all competitive baselines shown in~\cref{tab:pgd_eps4}. This result shows that ATAC does not exclusively rely on the directional information injected by the attacks, further demonstrating the effectiveness of our method even in extreme scenarios.

\begin{table*}[t]
\centering
\setlength{\tabcolsep}{4.5pt}
\renewcommand{\arraystretch}{1.2}
\resizebox{\textwidth}{!}{%
\begin{tabular}{l || cc || cc | cc || cc | cc }
\toprule
\multirow{3}{*}{\textbf{Dataset}} &
\multicolumn{6}{c||}{Robust accuracy (\%)} &
\multicolumn{4}{c}{Running time (s)} \\
\cline{2-11}
&
\multicolumn{2}{c||}{PGD $\epsilon=\sfrac{4}{255}$} &
\multicolumn{2}{c|}{Lure $\epsilon=\sfrac{4}{255}$} &
\multicolumn{2}{c||}{Avoid $\epsilon=\sfrac{4}{255}$} &
\multicolumn{2}{c|}{Lure $\epsilon=\sfrac{4}{255}$} &
\multicolumn{2}{c}{Avoid $\epsilon=\sfrac{4}{255}$} \\
& 
TTC & ATAC &
TTC & ATAC &
TTC & ATAC &
TTC & ATAC &
TTC & ATAC \\
\midrule
\textbf{CIFAR-100} & 
9.06 & 86.27 &
0.83 (\textcolor{blue}{-8.23}) & 5.83 (\textcolor{blue}{-80.44}) & 
1.67 (\textcolor{blue}{-7.39}) & 11.25 (\textcolor{blue}{-75.02}) & 0.150 & 0.650 & 0.114 & 0.652
\\
\midrule
\textbf{Caltech256} &
27.25 & 90.86 &
4.58 (\textcolor{blue}{-22.67}) & 15.83 (\textcolor{blue}{-75.03}) & 
8.75 (\textcolor{blue}{-18.50}) & 47.50 (\textcolor{blue}{-43.36}) & 0.149 & 0.656 & 0.111 & 0.658 
\\
\midrule
\textbf{OxfordPets} &
24.64 & 87.46 &
0.00 (\textcolor{blue}{-24.64}) & 10.83 (\textcolor{blue}{-76.63}) &
1.67 (\textcolor{blue}{-22.97}) & 45.83 (\textcolor{blue}{-41.63}) & 0.149 & 0.651 & 0.114 & 0.655 
\\
\midrule
\textbf{StanfordCars} &
12.84 & 70.80 & 
0.00 (\textcolor{blue}{-12.84}) & 4.17 (\textcolor{blue}{-66.63}) &
0.00 (\textcolor{blue}{-12.84}) & 18.33 (\textcolor{blue}{-52.47}) &  0.147 & 0.652 & 0.122 & 0.652 
\\
\midrule
\textbf{EuroSAT} &
13.57 & 66.57 &
0.83 (\textcolor{blue}{-12.74}) & 6.25 (\textcolor{blue}{-60.32}) & 
1.67 (\textcolor{blue}{-11.90}) & 9.17 (\textcolor{blue}{-57.40}) & 0.138 & 0.649 & 0.114 & 0.650 
\\
\midrule
\textbf{DTD} &
11.40 & 76.06 &
0.42 (\textcolor{blue}{-10.98}) & 3.33 (\textcolor{blue}{-72.73}) &
0.83 (\textcolor{blue}{-10.57}) & 15.42 (\textcolor{blue}{-60.64}) & 0.136 & 0.647 & 0.114 & 0.650 
\\
\midrule
\midrule
\textbf{Avg.} &
16.46 & 76.67 &
1.28 (\textcolor{blue}{-15.18}) & 7.21 (\textcolor{blue}{-69.46}) &
2.78 (\textcolor{blue}{-13.68}) & 24.25 (\textcolor{blue}{-52.42}) & 0.145 & 0.651 & 0.115 & 0.653 
\\
\bottomrule
\end{tabular}
}
\caption{
Comparison of TTC and ATAC against their respective adaptive attacks on 6 datasets. We report robust accuracies of both defenses in \%, as well as the running times of each attack in seconds per sample. Values in \blue{blue} indicate the decrease in robust accuracy compared to the non-adaptive PGD baseline. %Each experiment is conducted on 240 samples drawn from each dataset due to expensive computational overhead.
}
\label{tab:adaptive}
\end{table*}

\section{Robustness Against Adaptive Attacks}

A proper evaluation of adaptive or test-time adversarial defense methods, especially ones that include non-differentiable components, must take \emph{adaptive attacks} into account~\cite{tramer2020adaptiveattacksadversarialexample,carlini2019evaluatingadversarialrobustness}. To this end, we design two adaptive attacks specifically tailored against ATAC. These attacks have full access to all components of our method, including the augmentations used, the gating threshold $\tau^*$, and the correction step size $\alpha$. Following the guiding principle of~\cite{tramer2020adaptiveattacksadversarialexample}, our attacks adapt to all non-differentiable aspects of the defense.

Due to space limitations, we only provide a high-level intuition behind our attacks, and give a detailed overview along with their pseudocodes in the Appendix (\cref{sec:aaalgo}).

\paragraph{Lure Adaptive Attack.} This attack jointly optimizes the adversarial perturbation so that (i) the latent drift vectors are aligned, thereby increasing $\tau$ and activating the correction mechanism, and (ii) the adversarial loss is maximized over the whole pipeline. To achieve the latter, we use Expectation over Transformation (EOT)~\cite{athalye2018synthesizingrobustadversarialexamples}.

\paragraph{Avoid Adaptive Attack.} 

Contrary to the Lure attack, this attack aims to avoid activating the correction mechanism by reducing $\tau$. In addition, it also uses EOT to jointly maximize the loss over the original CLIP model.

\paragraph{Comparison to TTC.} In order to compare our method with related baselines, we conduct experiments with adaptive attacks against TTC~\cite{xing2025clipstrongfightback}. We evaluate the adaptive attack proposed in their paper that aims to reduce the $L_2$ distance between the visual embeddings of the image and its counterattacked variant. We also implement another adaptive attack for TTC that aims to jointly maximize the adversarial loss and the $L_2$ distance between the visual embeddings of the attacked and counterattacked images. For ease of comparison, we dub the former the Lure strategy and the latter the Avoid strategy against TTC. 

\subsection{Results}
We evaluate both TTC and ATAC against their corresponding adaptive attacks, with a budget of $\epsilon=4/255$ and present the results in~\cref{tab:adaptive}. While both defenses lose most of their robustness against these attacks, the robust accuracy retained by ATAC is significantly higher. This shows that even in worst-case conditions, \emph{ATAC still achieves nontrivial robustness}, further underscoring the effectiveness of our method. Furthermore, adaptive attacks take significantly longer against ATAC than against TTC, which is an additional benefit of ATAC in real-life worst-case scenarios.

Interestingly, the Avoid strategies perform worse against both defenses. This is only surprising for ATAC, where avoiding the correction mechanism would benefit a well-crafted adversarial attack. However, as demonstrated in~\cref{sec:main_robust_surpass_clean}, the objective of increasing the loss in an untargeted manner is at odds with creating less consistent drifts that would prevent correction, which explains why this attack is more difficult. On the other hand, we hypothesize that the success behind the Lure strategy lies in the attack's ability to perturb images in a way that is less mitigated by the augmentations.

\section{Conclusions and Future Work}

In this paper, we introduce ATAC, a novel test-time adversarial defense method. Our method is based on empirical observations and key shortcomings of related work. Our approach fills a gap in test-time defense strategies by explicitly estimating a semantic recovery direction in CLIP's feature space, using the visual features of augmented views. 

Through a rigorous experimental analysis, we show that our method achieves state-of-the-art robustness on 13 classification benchmarks, beating the previous best methods by an average of nearly 50\% in robust accuracy. We further demonstrate that unsupervised, gradient-based attacks overly rely on true-label supervision, inducing consistent shifts in CLIP's feature space, which allow ATAC to estimate a consistent recovery direction. However, even when this flaw is eliminated, ATAC still achieves state-of-the-art robustness. Furthermore, our method achieves nontrivial robustness against adaptive attacks at a comparatively large computational cost for the attacker, further underscoring the usability of ATAC in worst-case and real-life scenarios.

As the combination of adversarial defenses for CLIP is gaining traction, we hope future works can explore the combination of ATAC with other test-time defenses and adversarial fine-tuning methods. Moreover, future work could extend the novel paradigm pioneered by ATAC: correcting adversarial samples in feature space rather than image space.

{
    \small
    \bibliographystyle{ieeenat_fullname}
    \bibliography{main}

@misc{carlini2019evaluatingadversarialrobustness,
      title={On Evaluating Adversarial Robustness}, 
      author={Nicholas Carlini and Anish Athalye and Nicolas Papernot and Wieland Brendel and Jonas Rauber and Dimitris Tsipras and Ian Goodfellow and Aleksander Madry and Alexey Kurakin},
      year={2019},
      eprint={1902.06705},
      archivePrefix={arXiv},
      primaryClass={cs.LG},
      url={https://arxiv.org/abs/1902.06705}, 
}

@inproceedings{xing2025clipstrongfightback,
  title={Clip is strong enough to fight back: Test-time counterattacks towards zero-shot adversarial robustness of clip},
  author={Xing, Songlong and Zhao, Zhengyu and Sebe, Nicu},
  booktitle={Proceedings of the Computer Vision and Pattern Recognition Conference},
  pages={15172--15182},
  year={2025}
}

@inproceedings{dahal2025embeddingshiftdissectionclip,
  title={Embedding Shift Dissection on CLIP: Effects of Augmentations on VLM's Representation Learning},
  author={Dahal, Ashim and Murad, Saydul Akbar and Rahimi, Nick},
  booktitle={Proceedings of the Computer Vision and Pattern Recognition Conference},
  pages={4814--4818},
  year={2025}
}

@article{ilyas2019adversarialexamplesbugsfeatures,
  title={Adversarial examples are not bugs, they are features},
  author={Ilyas, Andrew and Santurkar, Shibani and Tsipras, Dimitris and Engstrom, Logan and Tran, Brandon and Madry, Aleksander},
  journal={Advances in neural information processing systems},
  volume={32},
  year={2019}
}

@inproceedings{mao2021adversarialattacksreversiblenatural,
  title={Adversarial attacks are reversible with natural supervision},
  author={Mao, Chengzhi and Chiquier, Mia and Wang, Hao and Yang, Junfeng and Vondrick, Carl},
  booktitle={Proceedings of the IEEE/CVF International Conference on Computer Vision},
  pages={661--671},
  year={2021}
}

@misc{lindqvist2021delvingpixelsadversarialsamples,
      title={Delving into the pixels of adversarial samples}, 
      author={Blerta Lindqvist},
      year={2021},
      eprint={2106.10996},
      archivePrefix={arXiv},
      primaryClass={cs.CV},
      url={https://arxiv.org/abs/2106.10996}, 
}

@inproceedings{zeng2019adversarialattacksimagespace,
  title={Adversarial attacks beyond the image space},
  author={Zeng, Xiaohui and Liu, Chenxi and Wang, Yu-Siang and Qiu, Weichao and Xie, Lingxi and Tai, Yu-Wing and Tang, Chi-Keung and Yuille, Alan L},
  booktitle={Proceedings of the IEEE/CVF Conference on Computer Vision and Pattern Recognition},
  pages={4302--4311},
  year={2019}
}

@inproceedings{sheng2025rtptimprovingadversarialrobustness,
  title={R-TPT: Improving Adversarial Robustness of Vision-Language Models through Test-Time Prompt Tuning},
  author={Sheng, Lijun and Liang, Jian and Wang, Zilei and He, Ran},
  booktitle={Proceedings of the Computer Vision and Pattern Recognition Conference},
  pages={29958--29967},
  year={2025}
}

@article{Krizhevsky2009LearningML,
  title={Learning multiple layers of features from tiny images},
  author={Krizhevsky, Alex and Hinton, Geoffrey and others},
  year={2009},
  publisher={Toronto, ON, Canada}
}

@inproceedings{Coates2011AnAO,
  title={An Analysis of Single-Layer Networks in Unsupervised Feature Learning},
  author={Adam Coates and A. Ng and Honglak Lee},
  booktitle={International Conference on Artificial Intelligence and Statistics},
  year={2011},
  url={https://api.semanticscholar.org/CorpusID:308212}
}

@article{FeiFei2006OneshotLO,
  title={One-shot learning of object categories},
  author={Li Fei-Fei and Rob Fergus and Pietro Perona},
  journal={IEEE Transactions on Pattern Analysis and Machine Intelligence},
  year={2006},
  volume={28},
  pages={594-611},
  url={https://api.semanticscholar.org/CorpusID:6953475}
}

@techreport{Griffin2007Caltech256OC,
  title={Caltech-256 object category dataset},
  author={Griffin, Gregory and Holub, Alex and Perona, Pietro and others},
  year={2007},
  institution={Technical Report 7694, California Institute of Technology Pasadena}
}

@InProceedings{parkhi12a,
  author       = "Omkar M. Parkhi and Andrea Vedaldi and Andrew Zisserman and C. V. Jawahar",
  title        = "Cats and Dogs",
  booktitle    = "IEEE Conference on Computer Vision and Pattern Recognition",
  year         = "2012",
}

@article{le2015tiny,
  title={Tiny imagenet visual recognition challenge},
  author={Le, Yann and Yang, Xuan},
  journal={CS 231N},
  volume={7},
  number={7},
  pages={3},
  year={2015}
}

@article{Nilsback2008AutomatedFC,
  title={Automated Flower Classification over a Large Number of Classes},
  author={Maria-Elena Nilsback and Andrew Zisserman},
  journal={2008 Sixth Indian Conference on Computer Vision, Graphics \& Image Processing},
  year={2008},
  pages={722-729},
  url={https://api.semanticscholar.org/CorpusID:15193013}
}

@inproceedings{Bossard2014Food101M,
  title={Food-101 - Mining Discriminative Components with Random Forests},
  author={Lukas Bossard and Matthieu Guillaumin and Luc Van Gool},
  booktitle={European Conference on Computer Vision},
  year={2014},
  url={https://api.semanticscholar.org/CorpusID:12726540}
}

@article{Krause20133DOR,
  title={3D Object Representations for Fine-Grained Categorization},
  author={Jonathan Krause and Michael Stark and Jia Deng and Li Fei-Fei},
  journal={2013 IEEE International Conference on Computer Vision Workshops},
  year={2013},
  pages={554-561},
  url={https://api.semanticscholar.org/CorpusID:14342571}
}

@inproceedings{Radford2021LearningTV,
  title={Learning Transferable Visual Models From Natural Language Supervision},
  author={Alec Radford and Jong Wook Kim and Chris Hallacy and Aditya Ramesh and Gabriel Goh and Sandhini Agarwal and Girish Sastry and Amanda Askell and Pamela Mishkin and Jack Clark and Gretchen Krueger and Ilya Sutskever},
  booktitle={International Conference on Machine Learning},
  year={2021},
  url={https://api.semanticscholar.org/CorpusID:231591445}
}

@article{Maji2013FineGrainedVC,
  title={Fine-Grained Visual Classification of Aircraft},
  author={Subhransu Maji and Esa Rahtu and Juho Kannala and Matthew B. Blaschko and Andrea Vedaldi},
  journal={ArXiv},
  year={2013},
  volume={abs/1306.5151},
  url={https://api.semanticscholar.org/CorpusID:2118703}
}

@article{Helber2017EuroSATAN,
  title={EuroSAT: A Novel Dataset and Deep Learning Benchmark for Land Use and Land Cover Classification},
  author={Patrick Helber and Benjamin Bischke and Andreas R. Dengel and Damian Borth},
  journal={IEEE Journal of Selected Topics in Applied Earth Observations and Remote Sensing},
  year={2017},
  volume={12},
  pages={2217-2226},
  url={https://api.semanticscholar.org/CorpusID:11810992}
}

@article{Cimpoi2013DescribingTI,
  title={Describing Textures in the Wild},
  author={Mircea Cimpoi and Subhransu Maji and Iasonas Kokkinos and Sammy Mohamed and Andrea Vedaldi},
  journal={2014 IEEE Conference on Computer Vision and Pattern Recognition},
  year={2013},
  pages={3606-3613},
  url={https://api.semanticscholar.org/CorpusID:4309276}
}

@misc{goodfellow2015explainingharnessingadversarialexamples,
      title={Explaining and Harnessing Adversarial Examples}, 
      author={Ian J. Goodfellow and Jonathon Shlens and Christian Szegedy},
      year={2015},
      eprint={1412.6572},
      archivePrefix={arXiv},
      primaryClass={stat.ML},
      url={https://arxiv.org/abs/1412.6572}, 
}

@inproceedings{carlini2017evaluatingrobustnessneuralnetworks,
  title={Towards evaluating the robustness of neural networks},
  author={Carlini, Nicholas and Wagner, David},
  booktitle={2017 ieee symposium on security and privacy (sp)},
  pages={39--57},
  year={2017},
  organization={Ieee}
}

@misc{madry2019deeplearningmodelsresistant,
      title={Towards Deep Learning Models Resistant to Adversarial Attacks}, 
      author={Aleksander Madry and Aleksandar Makelov and Ludwig Schmidt and Dimitris Tsipras and Adrian Vladu},
      year={2019},
      eprint={1706.06083},
      archivePrefix={arXiv},
      primaryClass={stat.ML},
      url={https://arxiv.org/abs/1706.06083}, 
}

@misc{guo2018counteringadversarialimagesusing,
      title={Countering Adversarial Images using Input Transformations}, 
      author={Chuan Guo and Mayank Rana and Moustapha Cisse and Laurens van der Maaten},
      year={2018},
      eprint={1711.00117},
      archivePrefix={arXiv},
      primaryClass={cs.CV},
      url={https://arxiv.org/abs/1711.00117}, 
}

@INPROCEEDINGS{8578289,
  author={Liao, Fangzhou and Liang, Ming and Dong, Yinpeng and Pang, Tianyu and Hu, Xiaolin and Zhu, Jun},
  booktitle={2018 IEEE/CVF Conference on Computer Vision and Pattern Recognition}, 
  title={Defense Against Adversarial Attacks Using High-Level Representation Guided Denoiser}, 
  year={2018},
  volume={},
  number={},
  pages={1778-1787},
  doi={10.1109/CVPR.2018.00191}}

@inproceedings{nie2022diffusionmodelsadversarialpurification,
  title={Diffusion Models for Adversarial Purification},
  author={Nie, Weili and Guo, Brandon and Huang, Yujia and Xiao, Chaowei and Vahdat, Arash and Anandkumar, Animashree},
  booktitle={International Conference on Machine Learning},
  pages={16805--16827},
  year={2022},
  organization={PMLR}
}

@inproceedings{cohen2019certifiedadversarialrobustnessrandomized,
  title={Certified adversarial robustness via randomized smoothing},
  author={Cohen, Jeremy and Rosenfeld, Elan and Kolter, Zico},
  booktitle={international conference on machine learning},
  pages={1310--1320},
  year={2019},
  organization={PMLR}
}

@article{shu2022testtimeprompttuningzeroshot,
  title={Test-time prompt tuning for zero-shot generalization in vision-language models},
  author={Shu, Manli and Nie, Weili and Huang, De-An and Yu, Zhiding and Goldstein, Tom and Anandkumar, Anima and Xiao, Chaowei},
  journal={Advances in Neural Information Processing Systems},
  volume={35},
  pages={14274--14289},
  year={2022}
}

@misc{feng2023diversedataaugmentationdiffusions,
      title={Diverse Data Augmentation with Diffusions for Effective Test-time Prompt Tuning}, 
      author={Chun-Mei Feng and Kai Yu and Yong Liu and Salman Khan and Wangmeng Zuo},
      year={2023},
      eprint={2308.06038},
      archivePrefix={arXiv},
      primaryClass={cs.CV},
      url={https://arxiv.org/abs/2308.06038}, 
}

@article{hassan2024alignpromptstesttimeprompting,
  title={Align your prompts: Test-time prompting with distribution alignment for zero-shot generalization},
  author={Abdul Samadh, Jameel and Gani, Mohammad Hanan and Hussein, Noor and Khattak, Muhammad Uzair and Naseer, Muhammad Muzammal and Shahbaz Khan, Fahad and Khan, Salman H},
  journal={Advances in Neural Information Processing Systems},
  volume={36},
  pages={80396--80413},
  year={2023}
}

@article{gao2021clip,
  title={Clip-adapter: Better vision-language models with feature adapters},
  author={Gao, Peng and Geng, Shijie and Zhang, Renrui and Ma, Teli and Fang, Rongyao and Zhang, Yongfeng and Li, Hongsheng and Qiao, Yu},
  journal={International Journal of Computer Vision},
  volume={132},
  number={2},
  pages={581--595},
  year={2024},
  publisher={Springer}
}

@inproceedings{mao2023understanding,
  title={Understanding Zero-shot Adversarial Robustness for Large-Scale Models},
  author={Mao, Chengzhi and Geng, Scott and Yang, Junfeng and Wang, Xin and Vondrick, Carl},
  booktitle={The Eleventh International Conference on Learning Representations}
}

@article{zhou2024revisiting,
  title={Revisiting the Adversarial Robustness of Vision Language Models: a Multimodal Perspective},
  author={Zhou, Wanqi and Bai, Shuanghao and Zhao, Qibin and Chen, Badong},
  journal={CoRR},
  year={2024}
}

@inproceedings{wang2024tapttesttimeadversarialprompt,
  title={Tapt: Test-time adversarial prompt tuning for robust inference in vision-language models},
  author={Wang, Xin and Chen, Kai and Zhang, Jiaming and Chen, Jingjing and Ma, Xingjun},
  booktitle={Proceedings of the Computer Vision and Pattern Recognition Conference},
  pages={19910--19920},
  year={2025}
}

@inproceedings{andriushchenko2020squareattackqueryefficientblackbox,
  title={Square attack: a query-efficient black-box adversarial attack via random search},
  author={Andriushchenko, Maksym and Croce, Francesco and Flammarion, Nicolas and Hein, Matthias},
  booktitle={European conference on computer vision},
  pages={484--501},
  year={2020},
  organization={Springer}
}

@article{ma2025safetyscalecomprehensivesurvey,
  title={Safety at scale: A comprehensive survey of large model and agent safety},
  author={Ma, Xingjun and Gao, Yifeng and Wang, Yixu and Wang, Ruofan and Wang, Xin and Sun, Ye and Ding, Yifan and Xu, Hengyuan and Chen, Yunhao and Zhao, Yunhao and others},
  journal={Foundations and Trends{\textregistered} in Privacy and Security},
  volume={8},
  number={3-4},
  pages={254--469},
  year={2025},
  publisher={Now Publishers, Inc.}
}

@inproceedings{Madry2017TowardsDL,
  title={Towards Deep Learning Models Resistant to Adversarial Attacks},
  author={Madry, Aleksander and Makelov, Aleksandar and Schmidt, Ludwig and Tsipras, Dimitris and Vladu, Adrian},
  booktitle={International Conference on Learning Representations},
  year={2018}
}

@article{schlarmann2024robustclip,
    title={Robust CLIP: Unsupervised Adversarial Fine-Tuning of Vision Embeddings for Robust Large Vision-Language Models}, 
    author={Christian Schlarmann and Naman Deep Singh and Francesco Croce and Matthias Hein},
    year={2024},
    journal={ICML}
}

@inproceedings{tong2024testtimealignment,
  title={Test-time Alignment-Enhanced Adapter for Vision-Language Models},
  author={Tong, Baoshun and Song, Kaiyu and Lai, Hanjiang},
  booktitle={ICASSP 2025-2025 IEEE International Conference on Acoustics, Speech and Signal Processing (ICASSP)},
  pages={1--5},
  year={2025},
  organization={IEEE}
}

@inproceedings{wang2024pretrainedmodelguidedfinetuning,
  title={Pre-trained model guided fine-tuning for zero-shot adversarial robustness},
  author={Wang, Sibo and Zhang, Jie and Yuan, Zheng and Shan, Shiguang},
  booktitle={Proceedings of the IEEE/CVF conference on computer vision and pattern recognition},
  pages={24502--24511},
  year={2024}
}

@inproceedings{pérez2021enhancingadversarialrobustnesstesttime,
  title={Enhancing adversarial robustness via test-time transformation ensembling},
  author={P{\'e}rez, Juan C and Alfarra, Motasem and Jeanneret, Guillaume and Rueda, Laura and Thabet, Ali and Ghanem, Bernard and Arbel{\'a}ez, Pablo},
  booktitle={Proceedings of the IEEE/CVF International Conference on Computer Vision},
  pages={81--91},
  year={2021}
}

@article{Zhou2021LearningTP,
  title={Learning to Prompt for Vision-Language Models},
  author={Kaiyang Zhou and Jingkang Yang and Chen Change Loy and Ziwei Liu},
  journal={International Journal of Computer Vision},
  year={2021},
  volume={130},
  pages={2337 - 2348},
  url={https://api.semanticscholar.org/CorpusID:237386023}
}

@inproceedings{li2024promptwordboostadversarial,
  title={One prompt word is enough to boost adversarial robustness for pre-trained vision-language models},
  author={Li, Lin and Guan, Haoyan and Qiu, Jianing and Spratling, Michael},
  booktitle={Proceedings of the IEEE/CVF Conference on Computer Vision and Pattern Recognition},
  pages={24408--24419},
  year={2024}
}

@inproceedings{croce2020reliableevaluationadversarialrobustness,
  title={Reliable evaluation of adversarial robustness with an ensemble of diverse parameter-free attacks},
  author={Croce, Francesco and Hein, Matthias},
  booktitle={International conference on machine learning},
  pages={2206--2216},
  year={2020},
  organization={PMLR}
}

@inproceedings{athalye2018synthesizingrobustadversarialexamples,
  title={Synthesizing robust adversarial examples},
  author={Athalye, Anish and Engstrom, Logan and Ilyas, Andrew and Kwok, Kevin},
  booktitle={International conference on machine learning},
  pages={284--293},
  year={2018},
  organization={PMLR}
}

@article{tramer2020adaptiveattacksadversarialexample,
  title={On adaptive attacks to adversarial example defenses},
  author={Tramer, Florian and Carlini, Nicholas and Brendel, Wieland and Madry, Aleksander},
  journal={Advances in neural information processing systems},
  volume={33},
  pages={1633--1645},
  year={2020}
}

@inproceedings{zhangclipure,
  title={CLIPure: Purification in Latent Space via CLIP for Adversarially Robust Zero-Shot Classification},
  author={Zhang, Mingkun and Bi, Keping and Chen, Wei and Guo, Jiafeng and Cheng, Xueqi},
  booktitle={The Thirteenth International Conference on Learning Representations}
}

@misc{deng2025fptnoisedynamicsceneawarecounterattack,
      title={FPT-Noise: Dynamic Scene-Aware Counterattack for Test-Time Adversarial Defense in Vision-Language Models}, 
      author={Jia Deng and Jin Li and Zhenhua Zhao and Shaowei Wang},
      year={2025},
      eprint={2510.20856},
      archivePrefix={arXiv},
      primaryClass={cs.CR},
      url={https://arxiv.org/abs/2510.20856}, 
}

@article{liu2025self,
  title={Self-Calibrated Consistency can Fight Back for Adversarial Robustness in Vision-Language Models},
  author={Liu, Jiaxiang and Du, Jiawei and Liu, Xiao and Tiwari, Prayag and Xu, Mingkun},
  journal={arXiv preprint arXiv:2510.22785},
  year={2025}
}
}

% \input{cvpr_template/sec/X_suppl}

% \section{Appendix(not done)}
% \clearpage
% \setcounter{page}{1}
\maketitlesupplementary
%\section{Implementation details}
%\label{sec:appx_implementation}
%Which CLIP variant is used, what are the hyperparameters of baselines, etc.

% \input{new_sec/app_imple_detail}

\section{Results Against Other Attacks}
\label{sec:appx_other_attacks}
\begin{table*}[t]
\centering
\setlength{\tabcolsep}{3pt}
\renewcommand{\arraystretch}{1.15}
\begin{tabular}{ll||c|c|c|c||c|c|c|c}
\toprule
\textbf{Dataset} &  & \multicolumn{4}{c||}{No Defense} & \multicolumn{4}{c}{ATAC} \\
\cline{3-10}
& & 
$auto_1$ & $auto_4$ & $CW_1$ & $CW_4$ & 
$auto_1$ & $auto_4$ & $CW_1$ & $CW_4$ \\

\midrule

\textbf{CIFAR10}   
& Rob. & 0.01 & 0.01 & 0.79 & 0.00 
& 84.72 (\textcolor{red}{+84.71}) & 85.18 (\textcolor{red}{+85.17}) & 79.24 (\textcolor{red}{+78.45}) & \textbf{91.58} (\textcolor{red}{+91.58}) \\
& Acc. & \textbf{85.08} & 85.08 & 85.08 & 85.08 
& 81.04 (\textcolor{blue}{-4.04}) & 81.04 (\textcolor{blue}{-4.04}) & 81.04 (\textcolor{blue}{-4.04}) & 81.04 (\textcolor{blue}{-4.04}) \\
\midrule
\textbf{CIFAR100}  
& Rob. & 0.11 & 0.11 & 0.30 & 0.00 
& 56.77 (\textcolor{red}{+56.66}) & 57.49 (\textcolor{red}{+57.38}) & 53.75 (\textcolor{red}{+53.45}) & \textbf{78.08} (\textcolor{red}{+78.08}) \\
& Acc. & \textbf{57.20} & 57.20 & 57.20 & 57.20 
& 53.74 (\textcolor{blue}{-3.46}) & 53.74 (\textcolor{blue}{-3.46}) & 53.74 (\textcolor{blue}{-3.46}) & 53.74 (\textcolor{blue}{-3.46}) \\
\midrule
\textbf{STL10}      
& Rob. & 0.00 & 0.00 & 11.86 & 0.01 
& 96.26 (\textcolor{red}{+96.26}) & 96.39 (\textcolor{red}{+96.39}) & 90.42 (\textcolor{red}{+78.56}) & \textbf{98.01} (\textcolor{red}{+98.00}) \\
& Acc. & \textbf{96.42} & 96.42 & 96.42 & 96.42 
& 95.72 (\textcolor{blue}{-0.70}) & 95.72 (\textcolor{blue}{-0.70}) & 95.72 (\textcolor{blue}{-0.70}) & 95.72 (\textcolor{blue}{-0.70}) \\
\midrule                            
\textbf{Flowers102}
& Rob. & 0.02 & 0.02 & 1.51 & 0.00 
& 64.12 (\textcolor{red}{+64.10}) & 64.79 (\textcolor{red}{+64.77}) & 48.77 (\textcolor{red}{+47.26}) & \textbf{84.92} (\textcolor{red}{+84.92}) \\
& Acc. & \textbf{65.56} & 65.56 & 65.56 & 65.56 
& 65.34 (\textcolor{blue}{-0.22}) & 65.34 (\textcolor{blue}{-0.22}) & 65.34 (\textcolor{blue}{-0.22}) & 65.34 (\textcolor{blue}{-0.22}) \\
\midrule                            
\textbf{FGVCAircraft} 
& Rob. & 0.09 & 0.09 & 0.00 & 0.00 
& 15.06 (\textcolor{red}{+14.97}) & 17.52 (\textcolor{red}{+17.43}) & 19.53 (\textcolor{red}{+19.53}) & \textbf{54.10} (\textcolor{red}{+54.10}) \\
& Acc. & \textbf{20.16} & 20.16 & 20.16 & 20.16 
& 19.83 (\textcolor{blue}{-0.33}) & 19.83 (\textcolor{blue}{-0.33}) & 19.83 (\textcolor{blue}{-0.33}) & 19.83 (\textcolor{blue}{-0.33}) \\
\midrule                              
\textbf{DTD}        
& Rob. & 0.16 & 0.16 & 2.55 & 0.05 
& 38.94 (\textcolor{red}{+38.78}) & 40.00 (\textcolor{red}{+39.84}) & 39.79 (\textcolor{red}{+37.24}) & \textbf{64.73} (\textcolor{red}{+64.68}) \\
& Acc. & \textbf{40.11} & 40.11 & 40.11 & 40.11 
& 39.15 (\textcolor{blue}{-0.96}) & 39.15 (\textcolor{blue}{-0.96}) & 39.15 (\textcolor{blue}{-0.96}) & 39.15 (\textcolor{blue}{-0.96}) \\
\midrule                            
\textbf{Avg.}        
& Rob. & 0.07 & 0.07 & 2.84 & 0.01 
& 59.31 (\textcolor{red}{+59.24}) & 60.23 (\textcolor{red}{+60.16}) & 55.25 (\textcolor{red}{+52.41}) & \textbf{78.57} (\textcolor{red}{+78.56}) \\
& Acc. & \textbf{60.76} & 60.76 & 60.76 & 60.76 
& 59.14 (\textcolor{blue}{-1.62}) & 59.14 (\textcolor{blue}{-1.62}) & 59.14 (\textcolor{blue}{-1.62}) & 59.14 (\textcolor{blue}{-1.62}) \\                          
\bottomrule
\end{tabular}
\vspace{5pt}
\caption{ATAC under various attacks. Here, \textit{auto} denotes AutoAttack and \textit{CW} denotes the Carlini--Wagner attack. The subscript indicates the attack budget $\epsilon$, e.g., $auto_1$ corresponds to AutoAttack with $\epsilon=1/255$. For AutoAttack, we adopt the ``plus'' version, which integrates untargeted attacks (APGD-CE, APGD-DLR, FAB), targeted attacks (APGD-T, FAB-T), and a gradient-free attack (Square), thereby providing a comprehensive and reliable evaluation of adversarial robustness.}
\label{tab:more_attacks}
\end{table*}
\begin{table*}[htbp]
\centering
\scriptsize
\setlength{\tabcolsep}{5pt}
\renewcommand{\arraystretch}{1.5}

\resizebox{\textwidth}{!}{%
\begin{tabular}{
  c
  c |
  S[table-format=2.2(2)] |
  *{4}{S[table-format=2.2(2)]} ||
  *{4}{S[table-format=2.2(2)]} |
  S[table-format=+2.2, retain-explicit-plus]
}
\toprule
\multicolumn{2}{c}{(\%)} &
\multicolumn{1}{|c|}{\multirow{2}{*}{CLIP}} &
\multicolumn{4}{c||}{\textbf{Adversarial Finetuning}} &
\multicolumn{4}{c|}{\textbf{Test-time Defense}} &
\multicolumn{1}{c}{\multirow{2}{*}{$\Delta$}} \\
& & &
\multicolumn{1}{c}{CLIP-FT} &
\multicolumn{1}{c}{TeCoA} &
\multicolumn{1}{c}{PMG-AFT} &
\multicolumn{1}{c||}{FARE} &
\multicolumn{1}{c}{TTE} &
\multicolumn{1}{c}{TTC} &
\multicolumn{1}{c}{R-TPT} &
\multicolumn{1}{c|}{\textbf{ATAC (ours)}} &
\\
\midrule

% ---------------- CIFAR10 ----------------

\multirow{2}{*}{\textbf{CIFAR10}} &
Rob. &
\num{0.74} & \num{3.34} & \num{33.61} & \num{40.66} & \num{19.65} &
\num{41.35(6.14)} & \num{28.75(0.18)} & \underline{\num{70.80}} & \textbf{\num{81.03}} &
\textcolor{red}{\num{+66.36}} \\
& Acc. &
\num{85.12} & \textbf{\num{84.90}} & \num{64.61} & \num{70.69} & \num{74.44} &
\num{84.74(0.40)} & \num{81.18(0.07)} & \num{82.19} & \num{81.03} &
\textcolor{blue}{\num{-4.09}} \\
\midrule

% ---------------- CIFAR100 ----------------

\multirow{2}{*}{\textbf{CIFAR100}} &
Rob. &
\num{0.26} & \num{0.90} & \num{18.95} & \num{22.52} & \num{11.40} &
\num{20.06(4.03)} & \num{14.31(0.25)} & \underline{\num{43.85}} & \textbf{\num{64.24}} &
\textcolor{red}{\num{+63.98}} \\
& Acc. &
\num{57.14} & \textbf{\num{59.51}} & \num{35.96} & \num{40.32} & \num{46.67} &
\num{58.61(0.25)} & \num{56.34(0.20)} & \num{52.69} & \num{53.64} &
\textcolor{blue}{\num{-3.50}} \\
\midrule

% ---------------- STL10 ----------------

\multirow{2}{*}{\textbf{STL10}} &
Rob. &
\num{11.0} & \num{12.73} & \num{70.08} & \num{73.08} & \num{59.06} &
\num{78.48(3.83)} & \num{76.70(0.23)} & \textbf{\num{90.59}} & \underline{\num{90.41}} &
\textcolor{red}{\num{+79.41}} \\
& Acc. &
\num{96.40} & \num{94.49} & \num{87.40} & \num{88.56} & \num{91.72} &
\textbf{\num{96.26(0.04)}} & \num{95.85(0.04)} & \num{96.09} & \num{95.71} &
\textcolor{blue}{\num{-0.69}} \\
\midrule

% ---------------- Caltech101 ----------------

\multirow{2}{*}{\textbf{Caltech101}} &
Rob. &
\num{14.67} & \num{14.21} & \num{55.51} & \num{61.08} & \num{50.74} &
\num{67.56(3.88)} & \num{65.78(0.07)} & \textbf{\num{79.32}} & \underline{\num{72.41}} &
\textcolor{red}{\num{+57.74}} \\
& Acc. &
\num{85.66} & \num{83.63} & \num{71.68} & \num{75.45} & \num{80.95} &
\num{85.84(0.09)} & \num{86.53(0.07)} & \textbf{\num{86.62}} & \num{85.14} &
\textcolor{blue}{\num{-0.52}} \\
\midrule

% ---------------- Caltech256 ----------------

\multirow{2}{*}{\textbf{Caltech256}} &
Rob. &
\num{8.47} & \num{6.76} & \num{43.19} & \num{45.91} & \num{38.79} &
\num{60.09(4.03)} & \num{60.11(0.04)} & \underline{\num{67.51}} & \textbf{\num{68.02}} &
\textcolor{red}{\num{+59.55}} \\
& Acc. &
\num{81.72} & \num{78.53} & \num{61.14} & \num{62.24} & \num{73.32} &
\textbf{\num{82.49(0.08)}} & \num{79.66(0.04)} & \num{77.67} & \num{80.72} &
\textcolor{blue}{\num{-1.00}} \\
\midrule

% ---------------- OxfordPets ----------------

\multirow{2}{*}{\textbf{OxfordPets}} &
Rob. &
\num{1.04} & \num{2.10} & \num{38.35} & \num{41.18} & \num{31.07} &
\num{50.33(7.30)} & \num{57.87(0.15)} & \underline{\num{71.79}} & \textbf{\num{77.11}} &
\textcolor{red}{\num{+76.07}} \\
& Acc. &
\num{87.44} & \num{84.14} & \num{62.12} & \num{65.88} & \num{79.37} &
\textbf{\num{88.13(0.13)}} & \num{83.35(0.21)} & \num{84.46} & \num{87.30} &
\textcolor{blue}{\num{-0.14}} \\
\midrule

% ---------------- Flowers102 ----------------

\multirow{2}{*}{\textbf{Flowers102}} &
Rob. &
\num{1.14} & \num{0.54} & \num{21.94} & \num{23.43} & \num{17.14} &
\num{35.88(4.72)} & \num{39.14(0.28)} & \underline{\num{52.07}} & \textbf{\num{54.74}} &
\textcolor{red}{\num{+53.60}} \\
& Acc. &
\num{65.46} & \num{53.37} & \num{36.80} & \num{37.00} & \num{47.98} &
\num{65.18(0.22)} & \num{64.16(0.19)} & \num{62.92} & \textbf{\num{65.34}} &
\textcolor{blue}{\num{-0.12}} \\
\midrule

% ---------------- FGVCAircraft ----------------

\multirow{2}{*}{\textbf{FGVCAircraft}} &
Rob. &
\num{0.00} & \num{0.00} & \num{2.49} & \num{2.22} & \num{1.35} &
\num{6.23(1.37)} & \underline{\num{13.77(0.38)}} & \num{13.62} & \textbf{\num{23.37}} &
\textcolor{red}{\num{+23.37}} \\
& Acc. &
\num{20.10} & \num{14.04} & \num{5.31} & \num{5.55} & \num{10.86} &
\textbf{\num{20.19(0.36)}} & \num{18.00(0.16)} & \num{19.14} & \num{19.80} &
\textcolor{blue}{\num{-0.30}} \\
\midrule

% ---------------- StanfordCars ----------------

\multirow{2}{*}{\textbf{StanfordCars}} &
Rob. &
\num{0.02} & \num{0.06} & \num{8.76} & \num{11.65} & \num{6.75} &
\num{22.36(4.17)} & \underline{\num{33.01(0.07)}} & \textbf{\num{43.75}} & \num{27.12} &
\textcolor{red}{\num{+27.10}} \\
& Acc. &
\num{52.02} & \num{42.11} & \num{20.91} & \num{25.44} & \num{38.68} &
\textbf{\num{52.73(0.31)}} & \num{48.16(0.16)} & \num{61.75} & \num{51.45} &
\textcolor{blue}{\num{-0.57}} \\
\midrule

% ---------------- Country211 ----------------

\multirow{2}{*}{\textbf{Country211}} &
Rob. &
\num{0.04} & \num{0.03} & \num{1.78} & \num{2.12} & \num{0.85} &
\num{3.05(0.89)} & \num{7.09(0.04)} & \underline{\num{8.80}} & \textbf{\num{30.14}} &
\textcolor{red}{\num{+30.10}} \\
& Acc. &
\num{15.25} & \num{12.07} & \num{4.75} & \num{4.64} & \num{9.26} &
\num{14.66(0.16)} & \num{13.08(0.05)} & \num{13.40} & \textbf{\num{16.52}} &
\textcolor{red}{\num{+1.27}} \\
\midrule

% ---------------- Food101 ----------------

\multirow{2}{*}{\textbf{Food101}} &
Rob. &
\num{0.70} & \num{0.42} & \num{13.90} & \num{18.57} & \num{11.65} &
\num{43.94(6.97)} & \num{57.84(0.15)} & \underline{\num{68.04}} & \textbf{\num{76.47}} &
\textcolor{red}{\num{+75.77}} \\
& Acc. &
\num{83.88} & \num{64.86} & \num{29.98} & \num{36.61} & \num{55.31} &
\textbf{\num{83.96(0.02)}} & \num{82.18(0.02)} & \num{83.41} & \num{83.57} &
\textcolor{blue}{\num{-0.31}} \\
\midrule

% ---------------- EuroSAT ----------------

\multirow{2}{*}{\textbf{EuroSAT}} &
Rob. &
\num{0.03} & \num{0.04} & \num{11.96} & \num{12.60} & \num{10.67} &
\num{6.91(2.13)} & \num{12.19(0.24)} & \underline{\num{14.16}} & \textbf{\num{66.90}} &
\textcolor{red}{\num{+66.87}} \\
& Acc. &
\num{42.59} & \num{27.64} & \num{16.58} & \num{18.53} & \num{21.88} &
\num{44.38(1.60)} & \num{53.24(0.09)} & \num{21.83} & \num{38.32} &
\textcolor{blue}{\num{-4.21}} \\
\midrule

% ---------------- DTD ----------------

\multirow{2}{*}{\textbf{DTD}} &
Rob. &
\num{2.98} & \num{2.39} & \num{17.61} & \num{14.95} & \num{15.64} &
\num{23.90(2.34)} & \num{27.32(0.25)} & \underline{\num{34.10}} & \textbf{\num{47.93}} &
\textcolor{red}{\num{+44.95}} \\
& Acc. &
\num{40.64} & \num{36.49} & \num{25.16} & \num{21.76} & \num{32.07} &
\textbf{\num{41.33(0.32)}} & \num{36.98(0.21)} & \num{42.66} & \num{39.15} &
\textcolor{blue}{\num{-1.49}} \\
\midrule

% ---------------- Avg ----------------

\multirow{2}{*}{\textbf{Avg.}} &
Rob. &
\num{3.16} & \num{3.35} & \num{26.01} & \num{28.46} & \num{21.14} &
\num{35.40} & \num{37.99} & \underline{\num{50.65}} & \textbf{\num{59.99}} &
\textcolor{red}{\num{+56.83}} \\
& Acc. &
\num{62.57} & \num{56.60} & \num{40.18} & \num{42.51} & \num{50.96} &
\textbf{\num{62.96}} & \num{61.44} & \num{60.37} & \num{61.37} &
\textcolor{blue}{\num{-1.20}} \\
\bottomrule
\end{tabular}
}% end resizebox

\caption{Classification accuracy (\%) on both adversarial images (Rob.) under 10-step PGD attack at $\epsilon_a = 1/255$ and clean images (Acc.) across datasets. Finetuning-based models are implemented as references. For test-time methods, we report mean$\pm$std over 3 runs.}
\label{tab:pgd_eps1_new}
\end{table*}

To further validate the generality and reliability of ATAC, we extend our evaluation beyond the standard PGD setting ($\epsilon=4/255$) to two widely recognized and complementary benchmarks: AutoAttack~\cite{croce2020reliableevaluationadversarialrobustness} and the Carlini–Wagner (CW) attack~\cite{carlini2017evaluatingrobustnessneuralnetworks}.
We use the ``plus'' version of AutoAttack that integrates six attacks, including both targeted and untargeted, as well as gradient-based and gradient-free attacks, in order to provide a standardized and rigorous robustness evaluation.  
% AutoAttack (``plus'' version) integrates several strong and diverse components—both gradient-based and gradient-free—to provide a standardized and rigorous robustness evaluation.
In contrast, the CW attack formulates adversarial example generation as an explicit optimization problem that seeks minimal perturbations leading to confident misclassification, making it a representative test of fine-grained vulnerability beyond gradient-based methods.
We evaluate both AutoAttack and CW under two perturbation budgets, $\epsilon \in {1/255, 4/255}$, to examine robustness under both mild and strong attack regimes. We further evaluate PGD with a budget of $\epsilon=1/255$.

%As summarized in Table~\ref{tab:more_attacks}, \textbf{ATAC delivers remarkable improvements in robust accuracy across all datasets and attack settings}. For example, on CIFAR10 under the CW attack with $\epsilon=4/255$, ATAC achieves a robust accuracy of \textbf{91.58\%}, compared to nearly $0\%$ without defense. Averaged over six datasets, ATAC improves robustness from \textbf{0.07\% / 2.84\%} (AutoAttack / CW at $\epsilon=1/255$) to \textbf{59.31\% / 55.25\%}, and from \textbf{0.01\% / 0.01\%} (AutoAttack / CW at $\epsilon=4/255$) to \textbf{60.23\% / 78.57\%}. Notably, these large gains come at only a minor cost in clean accuracy (average drop of \textbf{1.62\%}).

As shown in Tables~\ref{tab:more_attacks} and \ref{tab:pgd_eps1_new}, ATAC consistently achieves large gains in robust accuracy across all datasets and attack settings, while maintaining nearly unchanged clean performance. Even under strong attacks such as CW or AutoAttack at higher $\epsilon$, ATAC restores model predictions to a level comparable to or exceeding the clean baseline, highlighting its ability to generalize beyond PGD and effectively counter diverse adversaries.

Overall, these results confirm that \textbf{ATAC is not attack-specific}: it maintains strong and consistent robustness under a wide range of threat models, demonstrating its potential as a general-purpose test-time defense mechanism.
%\section{Results Against Other Attacks}
%\label{sec:appx_other_attacks}
%Self-explanatory: introduce the attacks and show the results.

\section{On the Distribution of Consistency-Scores} 
\label{sec:appx_tau_dist}
\begin{figure*}[htbp]
    \centering
    \includegraphics[width=1\linewidth]{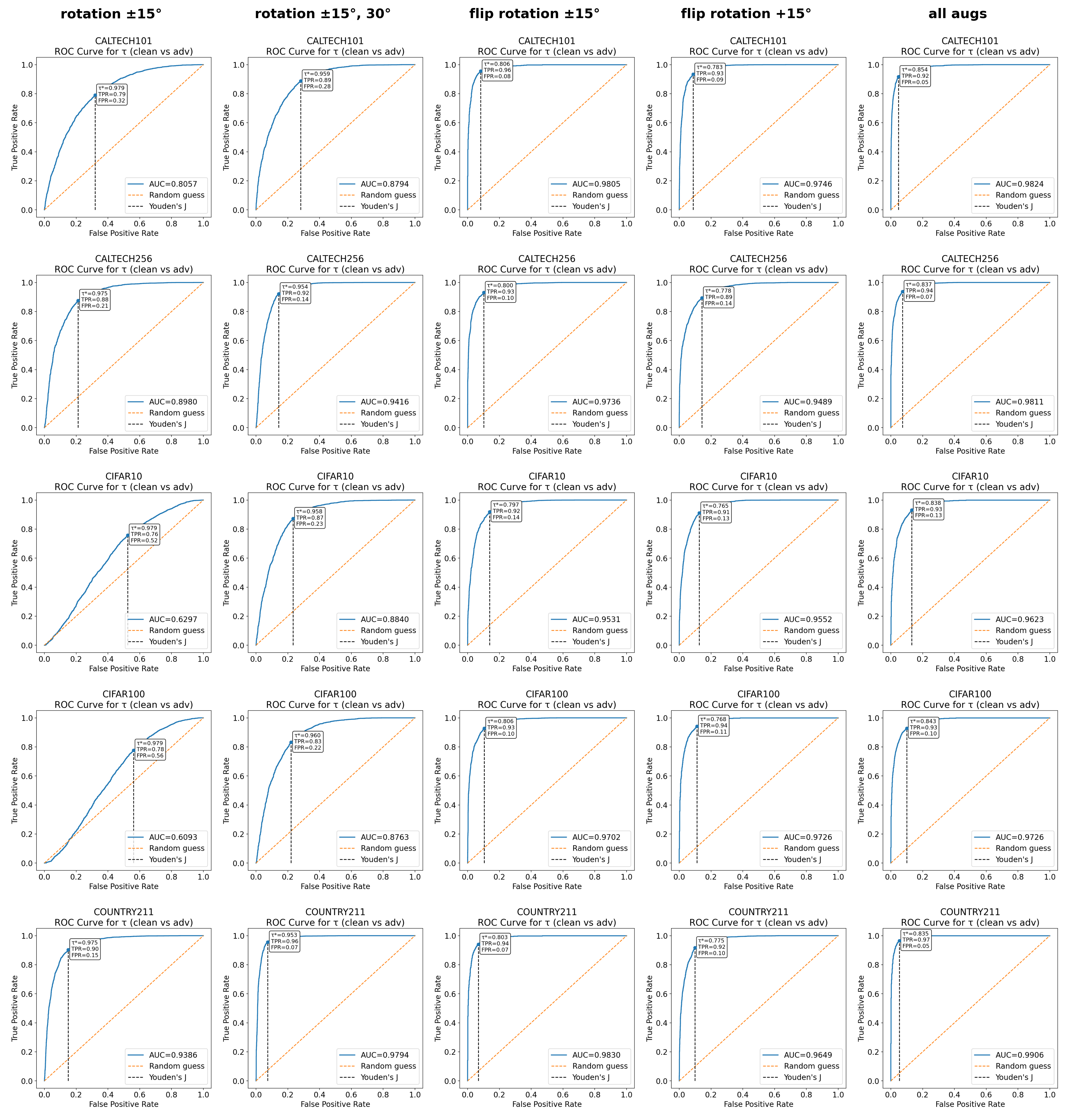}
    \caption{ROC curves of $\tau$-scores of different augmentation settings on different datasets.
    } 
    \label{fig:roc_augs}
\end{figure*}

In~\cref{sec:atac_intro} we argue that the augmentation-induced latent drift vectors are scattered for clean samples and consistent for adversarial inputs. To verify our claim, we analyze the distribution of $\tau$-scores for clean and adversarial inputs, and report the separability of the two distributions.

The last column of \cref{fig:roc_augs} shows the separability of clean and adversarial $\tau$-distributions using our set of augmentations. Our setting achieves a consistently high area under the curve (AUC) of nearly 1 in all cases, demonstrating that adversarial and clean inputs can be effectively separated using the consistency of their augmentation-induced latent drifts.

% We demonstrate in Fig.~\ref{fig:roc_augs} that the proposed $\tau$-scores effectively separate adversarial examples from clean ones. As shown, our augmentation configuration achieves the highest AUC among all compared settings, indicating the strongest discriminative capability.
%The figure that shows how the $\tau$-scores can separate adversarial examples from clean ones (ROC curves).

\section{Further Ablations}
\label{sec:appx_ablations}
\begin{table*}[t]
\centering
\setlength{\tabcolsep}{4pt}
\renewcommand{\arraystretch}{1.15}
\begin{tabular}{ll||c||c|c|c|c|c}
\toprule
\multirow{2}{*}{\textbf{Dataset}} & \multirow{2}{*}{} & \multirow{2}{*}{\textbf{No Defense}} & \multicolumn{5}{c}{\textbf{ATAC}} \\
\cline{4-8} % 只对ATAC的五列应用下划线
& & & $default$ & $asymmetric$ & $random$ & $color$ & $more$ \\
\midrule
\multirow{2}{*}{\textbf{STL10}}   
& Rob. & 0.00 & 97.94 (\textcolor{red}{+97.94}) & \textbf{98.00} (\textcolor{red}{+98.00}) & 41.06 (\textcolor{red}{+41.06}) & 2.69 (\textcolor{red}{+2.69}) & 81.94 (\textcolor{red}{+81.94}) \\
& Acc. & 96.19 & 95.38 (\textcolor{blue}{-0.81}) & 95.44 (\textcolor{blue}{-0.75}) & 95.81 (\textcolor{blue}{-0.38}) & 95.56 (\textcolor{blue}{-0.63}) & \textbf{96.19} (\textcolor{red}{+0.00}) \\
\midrule
\multirow{2}{*}{\textbf{Caltech101}} 
& Rob. & 0.00 & 67.63 (\textcolor{red}{+67.63}) & \textbf{67.81} (\textcolor{red}{+67.81}) & 31.13 (\textcolor{red}{+31.13}) & 6.25 (\textcolor{red}{+6.25}) & 55.94 (\textcolor{red}{+55.94}) \\
& Acc. & 68.38 & 67.69 (\textcolor{blue}{-0.69}) & 67.38 (\textcolor{blue}{-1.00}) & 68.31 (\textcolor{blue}{-0.07}) & \textbf{68.38} (\textcolor{red}{+0.00}) & \textbf{68.38} (\textcolor{red}{+0.00}) \\
\midrule
\multirow{2}{*}{\textbf{OxfordPets}}  
& Rob. & 0.00 & 95.56 (\textcolor{red}{+95.56}) & \textbf{95.88} (\textcolor{red}{+95.88}) & 42.25 (\textcolor{red}{+42.25}) & 2.44 (\textcolor{red}{+2.44}) & 84.88 (\textcolor{red}{+84.88}) \\
& Acc. & 83.13 & 83.25 (\textcolor{red}{+0.12}) & 83.13 (\textcolor{red}{+0.00}) & \textbf{83.31} (\textcolor{red}{+0.18}) & 82.81 (\textcolor{blue}{-0.32}) & 83.19 (\textcolor{red}{+0.06}) \\
\midrule
\multirow{2}{*}{\textbf{Flowers102}} 
& Rob. & 0.00 & \textbf{84.69} (\textcolor{red}{+84.69}) & 84.19 (\textcolor{red}{+84.19}) & 39.06 (\textcolor{red}{+39.06}) & 2.13 (\textcolor{red}{+2.13}) & 81.25 (\textcolor{red}{+81.25}) \\
& Acc. & 65.19 & 64.81 (\textcolor{blue}{-0.38}) & 64.75 (\textcolor{blue}{-0.44}) & 64.94 (\textcolor{blue}{-0.25}) & 64.31 (\textcolor{blue}{-0.88}) & \textbf{65.13} (\textcolor{blue}{-0.06}) \\
\midrule
\multirow{2}{*}{\textbf{FGVCAircraft}} 
& Rob. & 0.00 & 37.31 (\textcolor{red}{+37.31}) & \textbf{37.38} (\textcolor{red}{+37.38}) & 15.88 (\textcolor{red}{+15.88}) & 0.19 (\textcolor{red}{+0.19}) & 29.81 (\textcolor{red}{+29.81}) \\
& Acc. & 13.94 & 13.44 (\textcolor{blue}{-0.50}) & 13.25 (\textcolor{blue}{-0.69}) & 13.69 (\textcolor{blue}{-0.25}) & 13.69 (\textcolor{blue}{-0.25}) & \textbf{13.81} (\textcolor{blue}{-0.13}) \\
\midrule
\multirow{2}{*}{Avg.} 
& Rob. & 0.00 & 76.63 (\textcolor{red}{+76.63}) & \textbf{76.65} (\textcolor{red}{+76.65}) & 33.08 (\textcolor{red}{+33.08}) & 2.74 (\textcolor{red}{+2.74}) & 66.36 (\textcolor{red}{+66.36}) \\
& Acc. & 65.37 & 64.91 (\textcolor{blue}{-0.46}) & 64.79 (\textcolor{blue}{-0.58}) & 65.21 (\textcolor{blue}{-0.16}) & 64.95 (\textcolor{blue}{-0.42}) & \textbf{65.34} (\textcolor{blue}{-0.03}) \\
\bottomrule
\end{tabular}
\vspace{5pt}
\caption{Performance of ATAC with different augmentation settings under a 10-step PGD attack with $\epsilon = 4/255$, evaluated on 1,600 randomly sampled images from 5 datasets for each augmentation setting.}
\label{tab:augs}
\end{table*}
In~\cref{sec:ablations}, we find that the effect of $\alpha$ is minimal while $\tau^*$ is crucial. In this section, we investigate the effect of different augmentation choices. To understand which aspects of augmentations contribute to performance, we construct five ablation settings.

\begin{itemize}
    \item \textit{default}: the original setting used in our main experiments.  
    \item \textit{asymmetric}: when initially selecting augmentations, we hypothesized that averaging drift vectors of symmetric augmentations could reduce introduced bias. This setting is used to validate that hypothesis. % ????
    The augmentations in this setting are horizontal flip, and rotations with degrees +15, -20, -25, +30.
    \item \textit{random}: we replace the deterministic augmentations (horizontal flip with $p=1$ and fixed-degree rotations) in the default setting with random flips and rotations with a probability of $p=0.5$.  
    \item \textit{color}: replaces flip and rotation with color jittering augmentations. We used five random color jittering transformations, with brightness $\pm 40$, contrast $\pm 40$, saturation $\pm 40$, and hue $\pm 15$.
    \item \textit{more}: we include both horizontal and vertical flips, as well as 8 different rotations with degrees $\pm 15$, $\pm 20$, $\pm 25$, and $\pm 30$.
    %includes both horizontal and vertical flips, and introduces additional sets of rotations.  
\end{itemize}

% The exact configurations are provided in Appendix~\ref{sec:augs_settings}.  
As shown in Fig.~\ref{tab:augs}, there is only a negligible difference between \textit{default} and \textit{asymmetric}, indicating that symmetry does not necessarily improve performance. Moreover, varying rotation degrees can even yield improvements, offering more flexibility for augmentation choices in deployment.  
The \textit{random} setting does not achieve robustness comparable to the first two settings, although a moderate gain still exists. We hypothesize this is due to insufficient augmentation; extending the range could mitigate this deficiency but would also introduce instability and potentially degrade performance. 
The \textit{color} setting yields the poorest performance, which is consistent with the finding in \cite{dahal2025embeddingshiftdissectionclip} that CLIP’s representations are most affected by noise addition, followed by color-variant transformations (including color jitter). This also suggests that ATAC relies on label-preserving augmentations, while those that introduce substantial embedding shifts (e.g. noise addition, blur, coarse dropout...\cite{dahal2025embeddingshiftdissectionclip}) may be less suitable.
Finally, although the \textit{more} setting attains the highest clean accuracy, it yields roughly 10\% lower robust accuracy compared to \textit{default}, indicating that simply adding more augmentations does not necessarily lead to consistent gains. 
In practice, this shows that ATAC does not need many costly augmentations in deployment, as a small number of transformations already delivers high performance.

% In practice, this implies that real-world deployment need not trade off augmentation quantity for inference cost, as a small number of augmentations already delivers more satisfactory performance.

%Ablations on the set of augmentations.

% \subsection{Augmentation Settings in Ablation Study}
% \label{sec:augs_settings}

% We experiment with the following augmentation settings:

% \begin{itemize}[leftmargin=10pt]
%   \item \textbf{\textit{default}}: 
%   RandomHorizontalFlip (p=1.0); FixedRotation (+15, $-15$, +30, $-30$). 
%   \emph{(Symmetric rotations with horizontal flip.)}
  
%   \item \textbf{\textit{asymmetric}}: 
%   RandomHorizontalFlip (p=1.0); FixedRotation (+15, $-20$, +30, $-25$). 
%   \emph{(Non-symmetric rotation angles.)}
  
%   \item \textbf{\textit{random}}: 
%   RandomHorizontalFlip (p=0.5); RandomRotation ($\pm15$, $\pm15$, $\pm30$, $\pm30$). 
%   \emph{(Stochastic rotations with varying magnitudes.)}
  
%   \item \textbf{\textit{color}}: 
%   Five repetitions of ColorJitter ($\pm0.4$, $\pm0.4$, $\pm0.4$, $\pm0.1$), 
%   where parameters correspond to brightness, contrast, saturation, and hue. 
%   \emph{(Pure appearance-based augmentations.)}
  
%   \item \textbf{\textit{more}}: 
%   RandomHorizontalFlip (p=1.0); RandomVerticalFlip (p=1.0); 
%   FixedRotation (+15, $-15$, +20, $-20$, +25, $-25$, +30, $-30$). 
%   \emph{(Stronger augmentation with both flips and denser rotations.)}
% \end{itemize}

\section{Adaptive Attack Algorithms}
\label{sec:aaalgo}

\begin{algorithm}[t]
\DontPrintSemicolon
\caption{Adaptive ATAC Attack}
\label{alg:adaptive_atac}
\SetKwInput{KwIn}{Input}\SetKwInput{KwOut}{Output}
\KwIn{image $x\in[0,1]^{C\times H\times W}$}
\myinput{label $y$}
\myinput{CLIP image encoder $E_I$}
\myinput{text embeddings $\{t_i\}_{i=1}^k$}
\myinput{attack budget $\epsilon$}
\myinput{attack step size $\gamma$}
\myinput{strategy weight $\lambda$}
\myinput{optimization steps $T$}
\myinput{gating temperature $\mathcal{K}$}
\myinput{ATAC augmentation functions $\{\mathcal{A}_i\}_{i=1}^n$}
\myinput{ATAC correction step size $\alpha$} 
\myinput{ATAC gating threshold $\tau^*$} 
\myinput{attack strategy $\text{strategy} \in\{\text{avoid},\text{lure}\}$}
\KwOut{Adversarial perturbation $\delta^\star$ with $\|\delta^\star\|_\infty\le\epsilon$.}

% \textbf{Prep:} $F_{\text{txt}}\leftarrow \operatorname{norm}\!\left(f^{\text{txt}}_{\theta}(T)\right)$.\;
% Initialize $\delta$ (random or zero), project to $\|\cdot\|_\infty\le\epsilon$.\;
$\delta \sim \mathcal{U}(-\epsilon,+\epsilon)$

\For{$t=1\ldots T$}{
  % $x_{\text{adv}}\leftarrow\textsc{Clip}(x+\delta)$.\;
  $x_a = x + \delta$ \; \;
  
  \tcp{ATAC}
  $f_x \leftarrow E_I(x_a)$ \;
  $x_1, ..., x_n \leftarrow \mathcal{A}_i(x_a), ..., \mathcal{A}_i(x_a)$ \;
  $f_{x_1}, ..., f_{x_n} \leftarrow E_I(x_1), ..., E_I(x_n)$ \;
  $d_1, ..., d_n \leftarrow f_x - f_{x_1}, ..., f_x - f_{x_n}$ \;
  $\bar{d} \leftarrow \frac{1}{n}\sum_{i=1}^n d_i$ \;
  $\tau \leftarrow \frac{1}{n}\sum_{i=1}^n \cos(d_i, \bar{d})$ \; \;

  \tcp{Soft correction}
  $g \leftarrow \sigma(\mathcal{K} \cdot (\tau - \tau^*))$ \;
  $f^* \leftarrow  f_x + \alpha \cdot g \cdot \bar{d}$ \; \;

  \tcp{Strategy-dependent update}
  \eIf{$\mathrm{stragety} = \mathrm{avoid}$}{
    $\text{logits} \leftarrow \text{pred}(f_x, \{t_i\}_{i=1}^k)$\;
    $l = \mathcal{L}(\text{logits}, y) - \lambda \cdot \tau$ \;
  }{
    $\text{logits} \leftarrow \text{pred}(f^*, \{t_i\}_{i=1}^k)$ \;
    $l = \mathcal{L}(\text{logits}, y) + \lambda \cdot \tau$ \;
  }

  $\delta \leftarrow \prod_S(\delta + \gamma\cdot\text{sign}(\nabla_{\delta}l))$
}
$\delta^* \leftarrow \delta$

  % \tcp{EOT: replicate the same image K times}
  % Construct $x_{0,j}=x_{\text{adv}}$ for $j=1\ldots K$.\;
  % Compute embeddings $f_{0,j}\leftarrow f^{\text{img}}_{\theta}(x_{0,j})$.\;
  % Normalize each: $f_{0,j}\leftarrow\operatorname{norm}(f_{0,j})$.\;
  % Mean base embedding: $\bar{f_0}\leftarrow\frac{1}{K}\sum_{j=1}^K f_{0,j}$.\;

  % \tcp{Augmentation-induced drift}
  % For each of the $n$ fixed augmentation functions $A_i$:\;
  % \Indp Apply $A_i$ to each EOT copy: $x_{i,j}\leftarrow A_i(x_{0,j})$.\;
  % Compute $f_{i,j}=f^{\text{img}}_{\theta}(x_{i,j})$, normalize.\;
  % Per-view drift: $d_i\leftarrow \frac{1}{K}\sum_{j=1}^K (f_{i,j}-f_{0,j})$.\;
  % \Indm
  % Mean drift: $\bar{d}\leftarrow\frac{1}{n}\sum_{i=1}^n d_i$.\;

  % \tcp{Consistency score $\tau$}
  % For each $i$: $c_i\leftarrow\cos(d_i,\bar{d})$.\;
  % $\tau\leftarrow\frac{1}{n}\sum_{i=1}^n c_i$.\;

  % \tcp{ATAC correction}
  % $f^*\leftarrow \bar{f_0} + \alpha\,\bar{d}\quad\text{if}\quad \tau > \tau^*$.\;

  % \eIf{strategy==\textsc{avoid}}{
  %   $Z\leftarrow \exp(\text{logit\_scale})\cdot \bar{f_0} F_{\text{txt}}^\top$
  % }{
  %   $Z\leftarrow \exp(\text{logit\_scale})\cdot f_{\text{corr}} F_{\text{txt}}^\top$
  % }

  % $\mathcal{L}_{\text{CE}}\leftarrow \text{CE}(Z,y)$.\;

  % Add regularizer:\\
  % \eIf{strategy==\textsc{avoid}}{
  %   encourage $\tau\uparrow$
  % }{
  %   encourage $\tau\downarrow$
  % }
  
  % \tcp{Outer PGD update}
  % $\delta\leftarrow\delta + \eta\cdot\mathrm{sign}(\nabla_\delta\mathcal{L})$;\ 
  % project $\delta$ to $\|\delta\|_\infty\le\epsilon$, clamp $x+\delta$.\;
\Return $\delta^*$.
\end{algorithm}

\begin{algorithm}[t]
\DontPrintSemicolon
\caption{Adaptive TTC Attack}
\label{alg:adaptive_ttc}
\SetKwInput{KwIn}{Input}\SetKwInput{KwOut}{Output}
\KwIn{image $x\in[0,1]^{C\times H\times W}$}
\myinput{label $y$}
\myinput{CLIP image encoder $E_I$}
\myinput{text embeddings $\{t_i\}_{i=1}^k$}
\myinput{attack budget $\epsilon$}
\myinput{attack step size $\gamma$}
\myinput{strategy weight $\lambda$}
\myinput{optimization steps $T$}
\myinput{gating temperature $\mathcal{K}$}
\myinput{TTC counterattack budget $\epsilon_{ttc}$}
\myinput{TTC counterattack step size $\eta$}
\myinput{TTC gating threshold $\tau_{thresh}$}
\myinput{TTC noise budget $\epsilon_\tau$}
% \myinput{ATAC augmentation functions $\{\mathcal{A}_i\}_{i=1}^n$}
% \myinput{ATAC correction step size $\alpha$} 
% \myinput{ATAC gating threshold $\tau^*$} 
\myinput{attack strategy $\text{strategy} \in\{\text{avoid},\text{lure}\}$}
\KwOut{Adversarial perturbation $\delta^\star$ with $\|\delta^\star\|_\infty\le\epsilon$.}

$\delta \sim \mathcal{U}(-\epsilon, +\epsilon)$ \;
\For{$t = 1, ..., T$}{
    $x_a \leftarrow x + \delta$ \;
    $f_x \leftarrow E_I(x_a)$ \; \;
    \tcp{TTC step}
    $\delta_{ttc} \sim \mathcal{U}(-\epsilon, +\epsilon)$ \;
    $f_{x_{ttc}} \leftarrow E_I(x_a + \delta_{ttc})$ \;
    $\delta_{ttc} \leftarrow \prod_S(\delta_{ttc} + \eta \cdot \text{sign}(\nabla_{\delta_{ttc}} \lVert f_x - f_{x_{ttc}} \rVert_2))$ \; \;
    \tcp{Calculating $\hat{\tau}$ via EOT for thresholding}
    $\hat{\tau} \leftarrow 0$ \;
    \For{j = 1, ..., K}{
        $n_i \sim \mathcal{U}(-\epsilon_\tau, +\epsilon_\tau)$ \;
        $\hat{\tau} \leftarrow \hat{\tau} + \frac{1}{K} \cdot \frac{\lVert E_I(x_a + n) - f_x \rVert_2}{\lVert f_x\rVert_2}$\;
    }
    \;
    \tcp{Soft counterattack}
    $g \leftarrow \sigma(\mathcal{K} \cdot (\tau_{thresh} - \hat{\tau}))$ \;
    $x^* \leftarrow x_a + g\cdot\delta_{ttc}$ \; \;
    \tcp{Strategy-dependent update}
    \eIf{$\mathrm{strategy} = \mathrm{avoid}$}{
        $\text{logits} \leftarrow \text{pred}(f_x, \{t_i\}_{i=1}^k)$ \;
        $l \leftarrow \mathcal{L}(\text{logits}, y) + \lambda \cdot \hat{\tau}$
    }{
        $\text{logits} \leftarrow \text{pred}(E_I(x^*), \{t_i\}_{i=1}^k)$ \;
        $l \leftarrow \mathcal{L}(\text{logits}, y) - \lambda \cdot \hat{\tau}$
    }
    
$\delta \leftarrow \prod_S(\delta + \gamma\cdot\text{sign}(\nabla_{\delta}l))$
}
$\delta^* \leftarrow \delta$

\Return $\delta^*$.
\end{algorithm}

%\paragraph{ATAC}

Here, we give the full pseudocodes for our attacks. The adaptive attack against our method is given in~\cref{alg:adaptive_atac}, and the adaptive attack against TTC is given in~\cref{alg:adaptive_ttc}. In both pseudocodes, we use $\text{pred}(\cdot, \cdot)$ as a shorthand for the calculation of class-wise logits (see~\cref{eq:clip-classification}). $\sigma$ denotes the sigmoid function. As in the main text, we omit denoting the projection of adversarial attacks to the input space for the sake of simplicity.

For the adaptive attack against ATAC, we used $\epsilon=4/255$, $\gamma=1/255$, $\mathcal{K}=40$, $\lambda=1$, and $T=10$ optimization steps. 

Similarly, for the adaptive attack against TTC, we used $\epsilon=4/255$, $\gamma=1/255$, $\mathcal{K}=40$, $\lambda=1$, and $T=10$ optimization steps. The parameters of TTC were $\epsilon_{ttc}=2/255$, $\eta=1/255$, $\epsilon_\tau = 2/255$, and $\tau_{thresh}=0.2$.

In both cases, due to the large value of the gating temperature $\mathcal{K}$, the soft correcton and soft counterattack parts of our attacks can be interpreted as ``nearly hard'', and the hard variant would yield highly similar results.

\end{document}